\documentclass[11pt]{article}

\usepackage[preprint]{acl}

\usepackage{times}
\usepackage{latexsym}

\usepackage[T1]{fontenc}
\usepackage[utf8]{inputenc}

\usepackage{microtype}

\usepackage{inconsolata}

\usepackage{graphicx}

\usepackage{booktabs}
\usepackage{multirow}
\usepackage{relsize}
\usepackage[table]{xcolor}
\usepackage{arydshln}

\usepackage{amsmath, amssymb}

\usepackage[most]{tcolorbox}

\usepackage{enumitem}

\usepackage{caption}

\definecolor{revblue}{RGB}{0,64,192}
\newif\ifrevision
\revisionfalse
\ifrevision
  \newcommand{\rev}[1]{\textcolor{revblue}{#1}}
  \newenvironment{revblock}{\color{revblue}}{}
\else
  \newcommand{\rev}[1]{#1}
  \newenvironment{revblock}{}{}
\fi

\title{TIDE: Proactive Multi-Problem Discovery via Template-Guided Iteration}

\author{
    Soyeong Jeong$^{1}$ \; 
    Jinheon Baek$^{1}$ \;
    Minki Kang$^{1}$ \;
    Sung Ju Hwang$^{1,2}$ \\
    $^{1}$KAIST \;\; $^{2}$DeepAuto.ai \\
    \texttt{\{starsuzi, jinheon.baek, minkikang, sungju.hwang\}@kaist.ac.kr}}

\begin{document}
\maketitle

\begin{abstract}
Agents are widely deployed as assistants over documents, tools, and code. However, they typically act only on explicit user requests, which surface only the problems the user has noticed, while many other important problems coexist, hidden in plain sight, within the broader user context, with their total number unknown in advance. We frame this as the task of \textit{discovering multiple hidden problems from context}, in which coexisting problems should be uncovered, grounded in supporting evidence, and paired with concrete actions. To this end, we introduce \textsc{TIDE}, a template-guided iterative framework with two complementary mechanisms. Specifically, motivated by the observation that single-pass prediction anchors on the most salient cases and yields generic claims, we propose \textit{iterative discovery}, which surfaces a small batch of candidates per round while conditioning on what has already been found, so subsequent rounds extend coverage; and \textit{thought templates}, reusable schemas distilled from previously solved cases that specify what contextual signals to attend to and how to connect them, anchoring each prediction in a recognizable problem class. We validate \textsc{TIDE} on two realistic settings, personal workspaces and software repositories, across four model backbones, showing substantial gains over single-shot and parallel multi-agent baselines on task coverage, identification, and resolution.
\end{abstract}

\section{Introduction}

\begin{figure*}[t!]
    \centering
    \includegraphics[width=\linewidth]{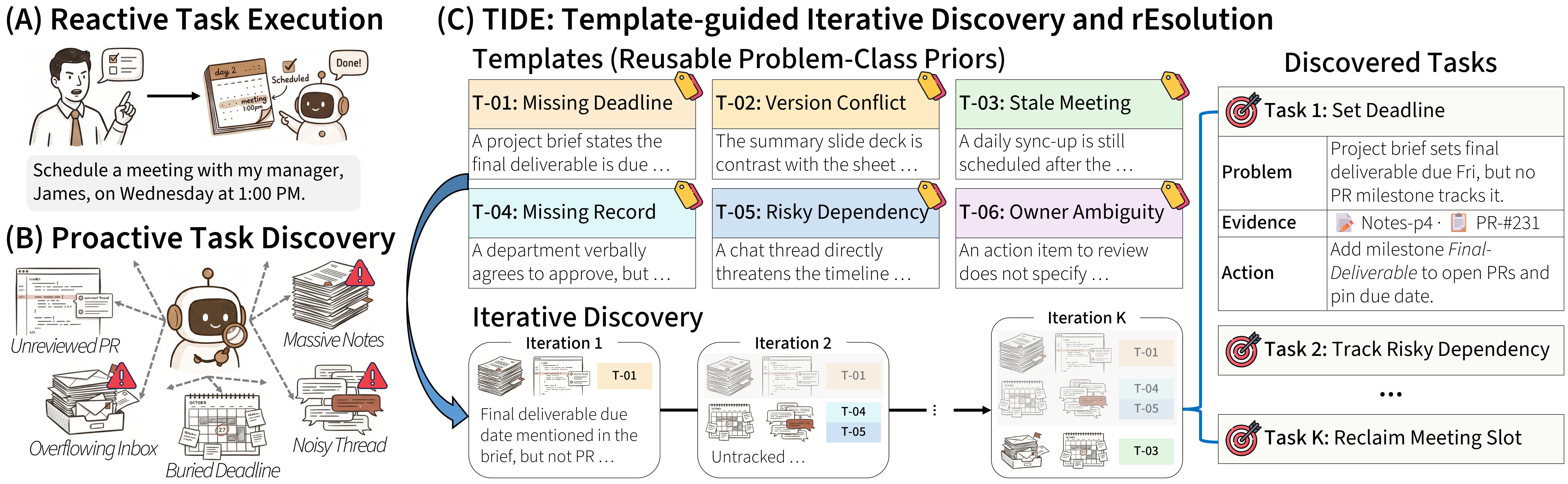}
    \vspace{-0.225in}
    \caption{Conceptual illustration of \textsc{TIDE}. (A) Reactive agents act only on explicit user requests, leaving (B) the many problems coexisting hidden across the user context untouched. (C) \textsc{TIDE} surfaces them by applying reusable thought templates over multiple rounds of iterative discovery conditioned on the cumulative state, returning per-task plans that identify, ground, and resolve each discovered problem.}
    \label{fig:concept}
    \vspace{-0.1in}
\end{figure*}

Large language model (LLM) agents are now routinely deployed as digital assistants that read documents, invoke external tools, and operate over complex environments~\cite{ReAct, SWE-agent, wu2024autogen}. Yet despite their growing capability, these agents remain reactive: they act only after a user issues an explicit request, whether to summarize a file, schedule a meeting, fix a failing test, or invoke a particular tool. This interaction model presumes that the user already knows what is wrong and what to ask.

In practice, however, the most consequential issues are often the ones a user has not yet noticed: a budget approval given verbally but not yet recorded in writing, holding up a vendor order on a hard deadline; two copies of the same report with conflicting numbers, both heading into an upcoming review; a recurring meeting the team has tacitly stopped attending, still blocking the only window for an urgent kickoff. Such issues sit, often in plain sight, inside the very documents, emails, and calendar entries that the agent could in principle inspect.

These otherwise different issues share a common structure that extends beyond the workspace setting above. Across the digital environments where agents operate (a personal workspace, a software repository, or another rich working context), evidence accumulates in which multiple problems coexist; none is articulated as a request; the number of coexisting problems is not known in advance; and resolving only the most salient one leaves the rest untouched. We therefore argue that proactive assistance is best framed not as anticipating a single user intent, but as the broader task of \textit{discovering multiple hidden problems from context}. Existing work on proactive agents has studied when to intervene~\cite{DBLP:conf/chi/LiuFSWIC25, DBLP:conf/emnlp/ZhangDLMKDCM25} or how to anticipate a single localized need~\cite{ProactiveAgent, ContextAgent, probe}, but largely sidesteps this multi-problem, context-wide setting that real workflows demand.

Meeting this task requires two complementary capabilities: broad coverage over coexisting problems that compete for attention with more salient ones, and enough precision per candidate to be actionable rather than speculative. To address these challenges, we present \textsc{TIDE} (\textbf{T}emplate-guided \textbf{I}terative \textbf{D}iscovery and r\textbf{E}solution; Figure~\ref{fig:concept}), a framework that combines two mechanisms operating along distinct axes. Specifically, motivated by the observation that single-pass prediction anchors on the most salient cases and yields generic claims, we propose \textit{iterative discovery}: rather than producing all predictions in one pass, the agent surfaces a small batch of candidates per round while conditioning on what has already been found, so that subsequent rounds are pushed beyond the cumulative discovery state. Additionally, for every surfaced candidate, the agent retrieves the artifacts that serve as evidence and commits to a concrete action that proposes a resolution, so that the per-case output forms a multi-element plan that simultaneously identifies, grounds, and addresses each discovered problem. We then complement iteration with \textit{thought templates}: reusable schemas distilled from previously solved cases, each capturing a recognizable pattern of hidden problem and laying out the chain of contextual signals from which that pattern can be inferred, anchoring each prediction in a known problem class instead of leaving the agent to infer it from scratch. Iteration thus expands the set of problems the agent will consider, while templates provide a reusable prior on how problems manifest in evidence, sharpening each prediction.

We instantiate this \textsc{TIDE} framework in two different real-world settings that share the underlying multi-problem structure: personal workspaces, where the agent surfaces multiple unresolved bottlenecks from workspace documents and proposes how to address each, and software repositories, where the agent identifies multiple hidden bugs directly from source code and produces patches that resolve them. Across both settings and four LLM backbones, \textsc{TIDE} consistently outperforms single-shot and parallel multi-agent baselines on task coverage, identification, and resolution; analyses further show that iterative discovery and thought templates contribute largely complementary gains and that templates transfer across backbones. Taken together, these results suggest that proactive assistance is better cast as an explicit, multi-step discovery process over context than as single-shot prediction from a user request, and they offer a general recipe for building agents that can both surface what users would not have thought to ask and point toward how to address it.

\section{Method}

Our method is motivated by three observations:
(1) the hidden problems within a given context are typically multiple in unknown number, and the most salient ones systematically overshadow subtler ones, so single-shot prediction leaves most problems undiscovered;
(2) generic prompting offers no reusable prior on how contextual signals become evidence for a particular class of problem, leading predictions to drift toward generic or speculative claims; and
(3) such patterns recur across instances and can be distilled from previously solved cases for reuse.
Guided by these observations, we first formalize the task (Section~\ref{sec:preliminaries}) and then describe our framework, \textsc{TIDE} (\textbf{T}emplate-guided \textbf{I}terative \textbf{D}iscovery and r\textbf{E}solution), which couples iterative discovery for broader coverage with thought templates for sharper fidelity (Section~\ref{sec:method_main}).

\subsection{Preliminaries: Hidden-Problem~Discovery~from~Context}
\label{sec:preliminaries}

\paragraph{Task Formulation}
We consider an agent operating over a collection of documents $\mathcal{D}$, broadly the digital artifacts available to the agent (e.g., emails and documents in a personal workspace, or functions in source code from a software repository). Within $\mathcal{D}$ there exists a latent set of \emph{hidden problems}, formalized as follows:
\[
\mathcal{P}^{\star} = \{p_1^{\star}, p_2^{\star}, \dots, p_n^{\star}\},
\]
where none of which is articulated as an explicit user request and whose cardinality $n$ is not known in advance. The objective is to produce a predicted set $\hat{\mathcal{P}}$ that approximates $\mathcal{P}^{\star}$, where each prediction takes the form of a triple $\hat{p} = (b, \hat{\mathcal{D}}, a)$ comprising a natural-language description $b$ of the candidate problem, a supporting subset $\hat{\mathcal{D}} \subseteq \mathcal{D}$ that grounds the prediction in evidence, and a concrete action $a$ that proposes a resolution. Solution quality therefore turns on two complementary axes: the \textit{coverage} over hidden problems in $\mathcal{P}^{\star}$, and the per-prediction \textit{fidelity}, with $b$ correctly describing the problem, $\hat{\mathcal{D}}$ providing valid grounding, and $a$ proposing an effective resolution.

\paragraph{Single-Shot Discovery}
Given this task, a natural baseline is to prompt a large language model to produce all problems in a single pass:
\[
\hat{\mathcal{P}} = \texttt{LLM}(\mathcal{D}).
\]
Yet, this formulation has two failure modes that correspond to the quality axes above: salient problems overshadow subtler ones, capping coverage; without any prior on what kinds of evidence patterns indicate a problem, predictions tend to drift toward generic or speculative claims, eroding fidelity.

\subsection{Template-Guided Iterative Discovery}
\label{sec:method_main}

To address these failure modes of single-shot discovery, we propose \textsc{TIDE} that couples two complementary components: \textit{thought templates}, reusable schemas distilled from previously solved cases that sharpen per-prediction fidelity, and \textit{iterative discovery}, which applies these templates over multiple rounds while conditioning on the cumulative state, broadening coverage over coexisting problems.

\paragraph{Thought Templates}
Rather than having the agent infer evidence patterns from scratch, we distill such patterns from prior cases into reusable discovery schemas, which we call thought templates. Specifically, each template specifies (i) a \emph{name} that labels a recurring class of hidden problem, (ii) a \emph{pattern} stating the structural form of that class, and (iii) an \emph{evidence flow}, an ordered sequence of contextual signals to attend to and how they should be connected to infer instances of that class.

Formally, let $\mathcal{T} = \{t_1, t_2, \dots, t_m\}$ denote the template set, where each template is a tuple:
\begin{equation}
t_i = (\text{name}_i,\ \text{pattern}_i,\ \text{evidence flow}_i).
\label{eq:template}
\end{equation}
Templates are constructed once from a collection of solved cases (in training instances) and held fixed at inference. To be more specific, for each solved case $\langle \mathcal{D}_{\texttt{train}}, p_{\texttt{train}}, r_{\texttt{train}} \rangle$ comprising the relevant document collection, the previously discovered problem, and a reference resolution (e.g., a patch in the software-repository setting), we prompt an LLM to abstract away instance-specific details and emit a template in the structured form of Equation~\ref{eq:template}:
$
t_i = \texttt{LLM}(\mathcal{D}_{\texttt{train}}, p_{\texttt{train}}, r_{\texttt{train}}).
$
At inference, the full set $\mathcal{T}$ is supplied to the agent as a library of discovery schemas, so that each prediction can be anchored in a recognizable problem class rather than inferred from scratch. An illustrative template from the workspace setting is below.

\begin{tcolorbox}[colback=gray!5, colframe=gray!50, boxrule=0.5pt, arc=2mm, left=4pt, right=4pt, top=3pt, bottom=3pt]
\noindent\textbf{Workspace Template:} Conflicting Source-of-Truth Blocks Sign-off Under Deadline\\[2pt]
\emph{Pattern:} A shared source artifact exists in conflicting versions across channels, and an imminent deadline blocks sign-off until one is made authoritative.\\
\emph{Evidence flow:} (i) locate the deliverable and its cited source; (ii) find conflicting copies across channels and confirm a material discrepancy; (iii) tie the conflict to a time-bounded review and the owner who can resolve it.
\end{tcolorbox}

\paragraph{Iterative Discovery and Resolution}
However, it is worth noting that even when equipped with such a template library, prompting the agent to produce all discoveries at once still concentrates its capacity on the most salient cases, leaving the subtler ones uncovered. To address this, we instead let the agent surface predictions over multiple rounds, with each round explicitly conditioned on what has already been found, so that subsequent rounds are pushed beyond the cumulative discovery state.

Formally, let $\hat{\mathcal{P}}^{(t)}$ denote the cumulative prediction state after round $t$, initialized as $\hat{\mathcal{P}}^{(0)} = \emptyset$. In round $t$, the agent generates a small batch of up to $k$ new candidate predictions in the triple form $(b, \hat{\mathcal{D}}, a)$ defined in Section~\ref{sec:preliminaries}, conditioned on the document collection $\mathcal{D}$, the template set $\mathcal{T}$, and the previous state $\hat{\mathcal{P}}^{(t-1)}$, formalized as follows:
\begin{equation}
\Delta \hat{\mathcal{P}}^{(t)} = \texttt{LLM}\!\left(\mathcal{D}, \mathcal{T}, \hat{\mathcal{P}}^{(t-1)}, k\right).
\label{eq:iterative_discovery}
\end{equation}
The state is then updated as $\hat{\mathcal{P}}^{(t)} = \hat{\mathcal{P}}^{(t-1)} \cup \Delta \hat{\mathcal{P}}^{(t)}$, and the loop terminates after $T$ rounds or earlier if a round returns empty, with the final output returned as $\hat{\mathcal{P}} = \hat{\mathcal{P}}^{(T)}$. We note that, by coupling identification with retrieval and a proposed action inside each per-prediction step, every round emits an actionable plan that simultaneously identifies, grounds, and addresses each surfaced problem, rather than treating these as separate downstream stages.

\section{Experimental Setup}
\label{sec:experimental_setup}
We evaluate \textsc{TIDE} on the multi-problem discovery task formalized in Section~\ref{sec:preliminaries}, whose goal is to surface multiple coexisting problems from a context $\mathcal{D}$ alone. Below, we describe our datasets, methods, evaluation metrics, and implementation details.

\subsection{Datasets}
We consider two real-world settings that share the underlying multi-problem structure, a personal workspace and a software repository; for each, we construct an evaluation split by extending existing data sources, since no existing benchmark directly targets multi-problem discovery from context.

\paragraph{Personal Workspace}
In this setting, each instance is the digital workspace of an individual user, consisting of a profile that captures the role, working style, current priorities, pain points, and relationships of that user, together with the workspace artifacts (documents, emails, and calendar entries) that constitute the context $\mathcal{D}$. Each problem is typically grounded in multiple workspace artifacts, so identifying it requires the agent to piece together evidence across several documents, emails, and calendar entries rather than reading it off any single artifact. The remaining artifacts in $\mathcal{D}$ act as distractors that look plausibly related to ongoing projects, relationships, and work, but are not implicated in any actual problem. A resolution takes the form of an action drawn from a predefined action set (such as sending an email, scheduling a meeting, sharing a document, or escalating to a manager), together with the parameters required to execute it (e.g., the recipients, subject, and body of an outgoing email). To instantiate this setting at scale, we build on the \rev{publicly released benchmark and} data construction pipeline of~\citet{probe}, \rev{which we extend only so that multiple problems coexist within a single workspace,} and produce $150$ problems across $30$ multi-problem workspaces, with $4$--$6$ problems and $88$--$113$ candidate artifacts per workspace.

\paragraph{Software Repository}
In this setting, each instance is a snapshot of a real-world open-source software repository at a commit where multiple unresolved bugs coexist and fixing them requires producing patches to multiple functions across the codebase. Each problem corresponds to an issue filed by a real GitHub user against the repository, and its gold resolution is the code patch from the pull request that fixed the issue. The context $\mathcal{D}$ is the set of candidate functions parsed from the snapshot; only a subset of these contains the coexisting bugs, while the rest are distractor functions that appear at the same snapshot but are not implicated in any bug. To instantiate this setting, we collect GitHub issues from Python repositories in \textsc{SWE-bench}~\cite{SWE-bench} and \textsc{TestExplora}~\cite{TestExplora}, and group same-repository issues at a common anchor commit at which the buggy functions of every grouped issue are unfixed; keeping only groups with at least two coexisting bugs spanning at least two buggy functions yields $146$ problems across $20$ multi-bug test instances drawn from $11$ projects, with $2$--$41$ problems and $6$--$646$ candidate functions per instance. \rev{The grouped bugs therefore coexisted in the repository's actual history, not through artificial pairing.}

\subsection{Methods}
We compare the following methods, all of which use the same backbone LLM (supporting the long-context) and operate over the same context $\mathcal{D}$, with the full $\mathcal{D}$ placed directly in the context window:

\begin{itemize}[leftmargin=*, itemsep=2pt, parsep=0pt, topsep=3pt]
    \item \textbf{\textsc{Single-Agent:}} Generates multiple problem predictions in a single LLM pass over $\mathcal{D}$.
    \item \textbf{\textsc{Multi-Agent:}} Runs multiple independent LLM agents in parallel over $\mathcal{D}$, matched in number to the rounds used by our iterative discovery.
    \item \rev{\textbf{\textsc{Self-Refine:}} Repeatedly critiques and rewrites an initial set of predictions~\cite{self-refine}, under the same LLM-call budget.}
    \item \rev{\textbf{\textsc{Reflexion:}} Follows the same refinement loop, but verbalizes a reflection on each attempt and keeps it in memory~\cite{reflexion}.}
    \item \textbf{\textsc{TIDE}} (Ours): Combines iterative discovery with thought templates, conditioning each round on the cumulative discovery state.
\end{itemize}

\begin{table*}[t!]
\small
\centering
\renewcommand{\arraystretch}{1.0}
\setlength{\tabcolsep}{2.5pt}
\resizebox{\linewidth}{!}{%
\begin{tabular}{l l cc cc cc cc cc cc >{\columncolor{gray!10}}c}
\toprule
& & \multicolumn{6}{c}{\textbf{Workspace}} & \multicolumn{6}{c}{\textbf{Repository}} & \multicolumn{1}{c}{} \\
\cmidrule(lr){3-8} \cmidrule(lr){9-14}
& & \multicolumn{2}{c}{\textbf{Retrieval}} & \multicolumn{2}{c}{\textbf{Identification}} & \multicolumn{2}{c}{\textbf{Resolution}}
  & \multicolumn{2}{c}{\textbf{Retrieval}} & \multicolumn{2}{c}{\textbf{Identification}} & \multicolumn{2}{c}{\textbf{Resolution}} & \multicolumn{1}{c}{} \\
\cmidrule(lr){3-4} \cmidrule(lr){5-6} \cmidrule(lr){7-8} \cmidrule(lr){9-10} \cmidrule(lr){11-12} \cmidrule(lr){13-14}
& \textbf{Methods}
  & \textbf{Cov.} & \textbf{F1} & \textbf{Cov.} & \textbf{F1} & \textbf{Cov.} & \textbf{F1}
  & \textbf{Cov.} & \textbf{F1} & \textbf{Cov.} & \textbf{F1} & \textbf{Cov.} & \textbf{F1}
  & \multicolumn{1}{c}{\textbf{Avg.}} \\
\midrule
\midrule
% ================= GPT =================
\multirow{5}{*}{\rotatebox{90}{\textbf{GPT}}}
& \textbf{\textsc{Single-Agent}}  & 47.60 & 54.32 & 47.85 & 54.63 & 49.67 & 56.14 & 8.66 & 10.34 & 11.15 & 12.92 & 12.19 & 13.27 & 31.56 \\
& \textbf{\textsc{Multi-Agent}}  & 32.15 & 45.41 & 27.24 & 38.85 & 29.49 & 41.62 & 10.11 & 12.66 & 10.19 & 12.77 & 9.89 & 12.39 & 23.56 \\
& \textbf{\textsc{Self-Refine}}  & 38.17 & 46.43 & 40.53 & 49.39 & 52.77 & 63.56 & 9.18 & 10.68 & 9.71 & 11.21 & 9.75 & 11.52 & 29.41 \\
& \textbf{\textsc{Reflexion}}  & 45.56 & 52.83 & 46.61 & 53.99 & 48.92 & 56.56 & 9.86 & 11.50 & 9.51 & 10.41 & 11.29 & 11.76 & 30.73 \\
& \textbf{\textsc{TIDE} (Ours)}  & \textbf{69.06} & \textbf{70.46} & \textbf{67.64} & \textbf{68.76} & \textbf{76.53} & \textbf{77.82} & \textbf{16.82} & \textbf{18.61} & \textbf{17.29} & \textbf{19.73} & \textbf{15.52} & \textbf{17.39} & \textbf{44.64} \\
\midrule
% ================= Gemini =================
\multirow{5}{*}{\rotatebox{90}{\textbf{Gemini}}}
& \textbf{\textsc{Single-Agent}}  & 50.01 & 61.48 & 42.99 & 52.95 & 35.95 & 44.31 & 13.08 & 17.20 & 13.34 & 16.90 & 15.08 & 18.71 & 31.83 \\
& \textbf{\textsc{Multi-Agent}}  & 46.10 & 58.54 & 38.98 & 49.95 & 32.44 & 41.54 & 14.75 & 18.51 & 14.71 & 17.72 & 15.32 & 18.89 & 30.62 \\
& \textbf{\textsc{Self-Refine}}  & 62.00 & 70.60 & 58.11 & 66.24 & 48.40 & 54.97 & 4.53 & 6.77 & 3.58 & 4.88 & 3.47 & 5.13 & 32.39 \\
& \textbf{\textsc{Reflexion}}  & 65.79 & 72.02 & 59.95 & 66.08 & 48.91 & 54.06 & 6.44 & 9.21 & 5.87 & 7.58 & 8.15 & 8.08 & 34.34 \\
& \textbf{\textsc{TIDE} (Ours)}  & \textbf{83.84} & \textbf{84.91} & \textbf{70.11} & \textbf{71.05} & \textbf{54.37} & \textbf{55.08} & \textbf{22.55} & \textbf{25.14} & \textbf{21.93} & \textbf{24.22} & \textbf{24.32} & \textbf{26.98} & \textbf{47.04} \\
\midrule
% ================= Claude =================
\multirow{5}{*}{\rotatebox{90}{\textbf{Claude}}}
& \textbf{\textsc{Single-Agent}}  & 13.82 & 20.48 & 13.53 & 17.16 & 11.79 & 14.67 & 12.85 & 16.09 & 9.86 & 12.18 & 9.84 & 12.34 & 13.72 \\
& \textbf{\textsc{Multi-Agent}}  & 21.60 & 28.89 & 18.04 & 24.02 & 15.99 & 20.69 & 9.37 & 13.09 & 7.96 & 10.98 & 8.70 & 10.79 & 15.84 \\
& \textbf{\textsc{Self-Refine}}  & 30.12 & 34.39 & 32.51 & 35.72 & 30.18 & 33.29 & 11.85 & 14.55 & 9.68 & 11.72 & 9.92 & 11.79 & 22.14 \\
& \textbf{\textsc{Reflexion}}  & 28.74 & 33.82 & 30.35 & 35.08 & 27.13 & 30.91 & 13.08 & 15.98 & 9.39 & 11.03 & 10.26 & 12.52 & 21.52 \\
& \textbf{\textsc{TIDE} (Ours)}  & \textbf{32.01} & \textbf{37.01} & \textbf{35.77} & \textbf{41.63} & \textbf{30.99} & \textbf{36.58} & \textbf{19.99} & \textbf{22.70} & \textbf{14.50} & \textbf{16.50} & \textbf{15.79} & \textbf{17.76} & \textbf{26.77} \\
\midrule
% ================= Qwen =================
\multirow{5}{*}{\rotatebox{90}{\textbf{Qwen}}}
& \textbf{\textsc{Single-Agent}}  & 30.46 & 39.90 & 32.44 & 42.37 & 27.13 & 35.67 & 5.60 & 6.49 & 5.34 & 6.29 & 4.59 & 5.47 & 20.15 \\
& \textbf{\textsc{Multi-Agent}}  & 39.34 & 52.12 & 31.04 & 42.27 & 26.21 & 35.37 & 5.00 & 6.72 & 6.58 & 7.50 & 5.83 & 6.50 & 22.04 \\
& \textbf{\textsc{Self-Refine}}  & 24.40 & 29.96 & 24.71 & 30.42 & 23.63 & 29.29 & 6.84 & 8.68 & 6.65 & 8.71 & 5.30 & 6.51 & 17.09 \\
& \textbf{\textsc{Reflexion}}  & 31.54 & 40.49 & 34.82 & 44.66 & 29.34 & 36.74 & 7.50 & 8.74 & 8.62 & 9.86 & 7.50 & 8.24 & 22.34 \\
& \textbf{\textsc{TIDE} (Ours)}  & \textbf{52.39} & \textbf{60.21} & \textbf{50.50} & \textbf{58.13} & \textbf{41.87} & \textbf{48.06} & \textbf{8.88} & \textbf{10.48} & \textbf{9.60} & \textbf{11.19} & \textbf{8.52} & \textbf{10.11} & \textbf{30.83} \\
\bottomrule
\end{tabular}
}
\vspace{-0.1in}
\caption{Main results on the Personal Workspace and Software Repository settings, reporting Coverage (Cov.) and F1 over three independent runs; the best per-LLM results are in \textbf{bold}.}
\label{tab:main}
\vspace{-0.1in}
\end{table*}

\subsection{Evaluation Metrics}
Since each instance contains multiple gold problems and the model surfaces multiple predictions, our metrics score individual gold-prediction pairs and aggregate them into instance-level scores.

\paragraph{Scoring Components}
Each matched (gold, prediction) pair is scored along three components: \textit{retrieval}, \textit{identification}, and \textit{resolution}. Specifically, retrieval is measured by the overlap between the predicted and gold-annotated evidence IDs, while identification and resolution are scored by an LLM judge~\cite{geval}\rev{\footnote{The judge is reliable: it matches the human label on $89.2\%$ of pairs, as often as the annotators match one another ($88.1\%$), and the ranking of methods is preserved under two cross-family judges (please see Appendix~\ref{appen:judge} for details).}} on a Likert-style rubric against the gold problem description and the gold reference action, respectively (See Appendix~\ref{appen:prompts}).

\paragraph{Coverage and F1}
To aggregate per-pair scores into instance-level metrics, we use retrieval as the matching score, pairing each gold problem with its best-scoring prediction and each prediction with its best-scoring gold. The coverage of a component is then the matched score averaged over all gold problems, capturing how well the agent discovers each of the hidden problems. The F1 of a component is the harmonic mean of this coverage and the analogous matched score averaged over predictions. \rev{This average keeps at most one prediction per gold, the highest-scoring among those paired with it, and leaves out predictions paired with no gold.} Both metrics are macro-averaged across instances.

\subsection{Implementation Details}
We instantiate the agent with each of four LLMs that support long-context: GPT-5 mini~\cite{openai2025gpt5}, Claude Sonnet 4.5~\cite{anthropic2025sonnet45}, Gemini 3.5 Flash~\cite{google2026gemini35flash}, and Qwen 3.6 Flash~\cite{qwen2026qwen36}. The LLM judge is fixed to GPT-5 mini. Thought templates are constructed by each LLM on its own from a held-out set of solved training cases (disjoint from the test split\rev{; see Appendix~\ref{appen:disjoint}}), yielding $20$ templates for the personal-workspace setting and $108$ templates for the software-repository setting. For iterative discovery, we set $T=10$ rounds for the personal-workspace setting and $T=3$ for the software-repository setting; in both cases, we condition each round on the cumulative state $\hat{\mathcal{P}}^{(t-1)}$ and terminate early when a round returns an empty batch.
\rev{We refer readers to Appendix~\ref{appen:impl} for further details.}

\begin{figure*}[t!]
    \centering
    \includegraphics[width=0.92\linewidth]{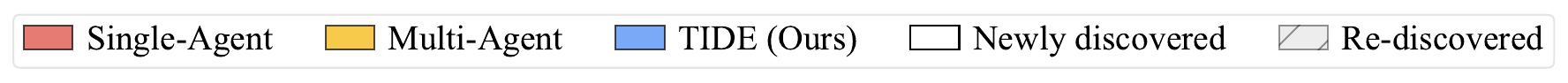}
    \\[-0.02in]
    \begin{minipage}[t]{0.62\linewidth}
        \centering
        \includegraphics[width=0.975\linewidth]{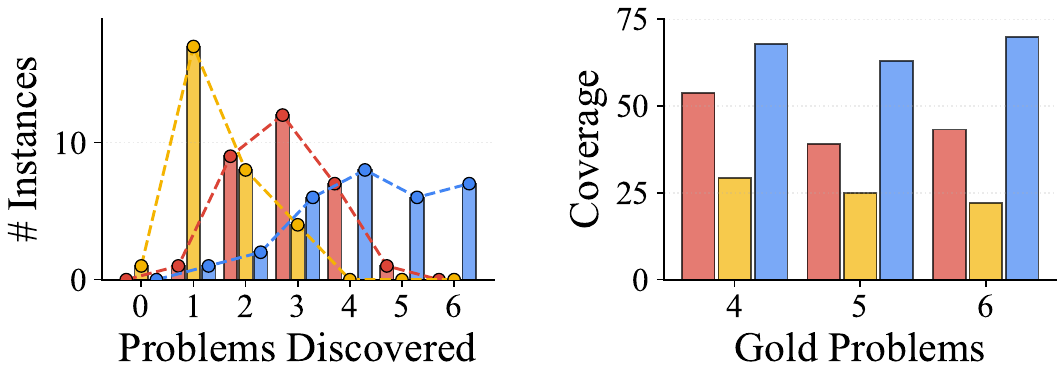}
        \vspace{-0.075in}
        \captionof{figure}{Multi-problem discovery on the Workspace setting with GPT. Left: discovered problems per instance. Right: coverage by gold count.}
        \label{fig:multi_bottleneck_scaling}
    \end{minipage}\hfill
    \begin{minipage}[t]{0.36\linewidth}
        \centering
        \includegraphics[width=0.975\linewidth]{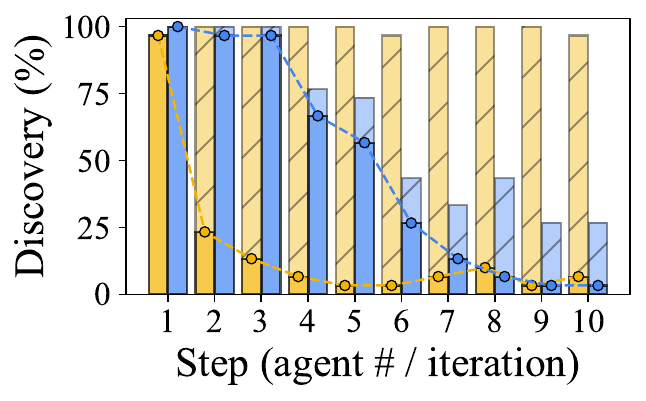}
        \vspace{-0.075in}
        \captionof{figure}{Newly vs.\ re-discovered predictions on the Workspace task with GPT.}
        \label{fig:diversity}
    \end{minipage}
\end{figure*}

\section{Results and Analyses}

\subsection{Main Results}
Table~\ref{tab:main} shows performance on retrieval, identification, and resolution across the Workspace and Repository settings, under four LLM backbones.
Although recent LLMs can process the entire candidate pool in a single long-context pass, this access alone is insufficient for multi-problem discovery.
This is reflected in the \textsc{Single-Agent} baseline, which commits to its first hypotheses without revisiting the context and leaves the majority of gold problems undiscovered.
Surprisingly, running the same backbone as multiple independent agents in parallel does not close this gap.
\rev{Allowing the agent to revise its own answer does not close the gap either, as the benefit of \textsc{Self-Refine} and \textsc{Reflexion} over \textsc{Single-Agent} appears on some backbones and reverses on others, which suggests that what matters is not iteration alone but conditioning each round on what has already been found.}
In contrast, \textsc{TIDE} combines iterative discovery with reusable templates, consistently achieving the best performance across retrieval, identification, and resolution.
\rev{\textsc{TIDE} holds this lead across all backbones and settings, and its gains are statistically significant in most comparisons (Appendix~\ref{appen:significance}).}

\subsection{In-Depth Analyses}

\begin{figure}[t!]
    \centering
    \includegraphics[width=0.975\linewidth]{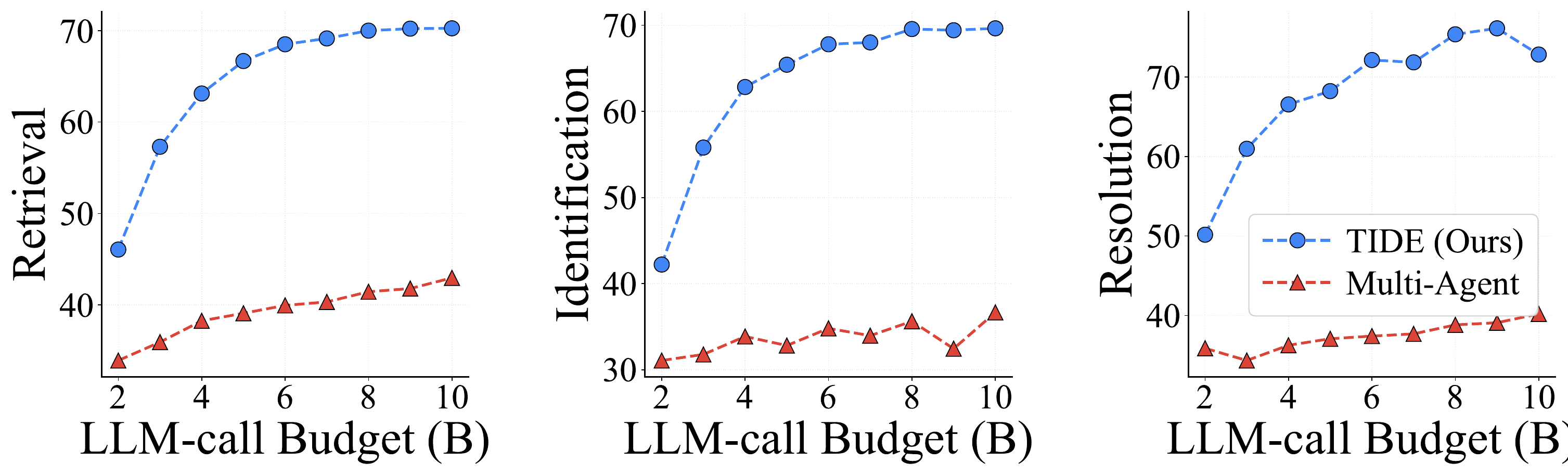}
    \vspace{-0.075in}
    \caption{F1 results as a function of the per-instance LLM-call budget \rev{$B$} on the Workspace setting with GPT.}
    \label{fig:budget_scaling}
    \vspace{-0.08in}
\end{figure}

\paragraph{Discovery on Multi-Problem Instances.}
Our framework targets multi-problem instances, where recovering a single problem is not enough.
To directly assess this capability, we report results in Figure~\ref{fig:multi_bottleneck_scaling}, where every instance contains four to six gold problems.
\rev{\textsc{TIDE} reaches four or more problems in most instances (left), which neither baseline does.}
Moreover, across instances with increasing numbers of gold problems (right), \textsc{TIDE} consistently continues to recover most of them while the baselines lag further behind, with \textsc{Multi-Agent} even falling below the simpler \textsc{Single-Agent}.

\paragraph{Effectiveness of Iterative Discovery.}
To understand why \textsc{Multi-Agent} fails to match \textsc{TIDE} under the same LLM-call budget, we decompose each step's predictions into two categories: newly discovered and re-discovered.
As shown in Figure~\ref{fig:diversity}, both methods start from the same point at the first step, but from the second step onward, \textsc{Multi-Agent}'s newly discovered items drop sharply and re-discovered items take over the majority of its predictions.
\textsc{TIDE}, by contrast, keeps contributing newly discovered problems across the following steps.
Each agent in \textsc{Multi-Agent} runs without access to what others have surfaced, so it re-anchors on the same most-salient signal.
\textsc{TIDE}, instead, conditions each step on the cumulative discovery state, redirecting capacity toward additional problems that independent parallel agents leave uncovered and driving its coverage lead in Table~\ref{tab:main}.
\rev{Even with deduplication or diversity prompting, \textsc{Multi-Agent} cannot recover what no agent found (Appendix~\ref{appen:variants}).}

\paragraph{Effect of LLM-Call Budget.}
The diversity analysis explains why \textsc{Multi-Agent} stops accumulating new problems, but it leaves open whether a larger budget could close the gap.
To address this, we vary the per-instance LLM-call budget \rev{$B$} from 2 to 10, where \rev{$B$} corresponds to the iteration cutoff for \textsc{TIDE} and to the number of aggregated parallel agents for \textsc{Multi-Agent}.
As shown in Figure~\ref{fig:budget_scaling}, \textsc{TIDE} scales steeply with \rev{$B$}, while \textsc{Multi-Agent} plateaus early. Interestingly, \textsc{Multi-Agent} at \rev{$B{=}10$} still falls below \textsc{TIDE} at \rev{$B{=}2$}.
This suggests that the gains of iterative discovery extend beyond retrieval coverage to identification and resolution, and more fundamentally, that scaling parallel agents is no substitute for iterative conditioning.
\rev{We further report the tokens and cost of each method under this budget (Appendix~\ref{appen:token}).}

\begin{figure}[t!]
    \centering
    \includegraphics[width=0.975\linewidth]{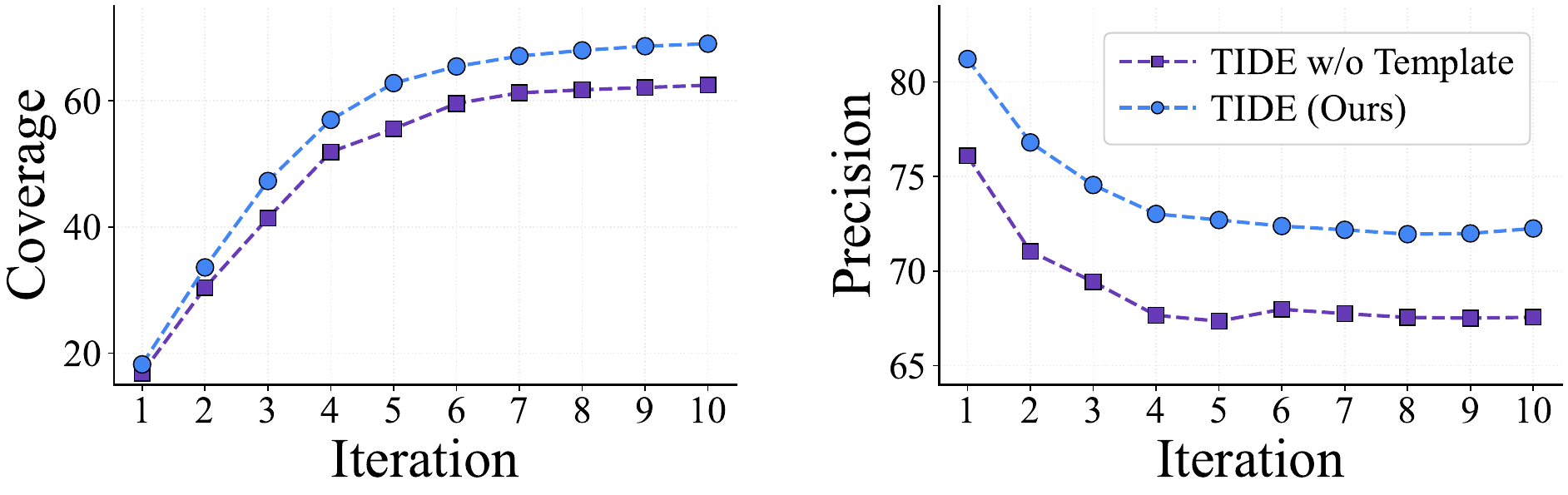}
    \vspace{-0.075in}
    \caption{Per-iteration retrieval coverage (left) and precision (right) on the Workspace setting with GPT.}
    \label{fig:iter_recall_precision}
    \vspace{-0.05in}
\end{figure}

\begin{table}[t!]
\small
\centering
\renewcommand{\arraystretch}{1.05}
\renewcommand{\tabcolsep}{0.9mm}
\resizebox{\linewidth}{!}{%
\begin{tabular}{l l cc cc cc}
\toprule
 & & \multicolumn{2}{c}{\textbf{Retrieval}}
 & \multicolumn{2}{c}{\textbf{Identification}}
 & \multicolumn{2}{c}{\textbf{Resolution}} \\
\cmidrule(lr){3-4}\cmidrule(lr){5-6}\cmidrule(lr){7-8}
 & \textbf{Method} & \textbf{Cov.} & \textbf{F1} & \textbf{Cov.} & \textbf{F1} & \textbf{Cov.} & \textbf{F1} \\
\midrule
\midrule
\multirow{3}{*}{\rotatebox{90}{\textbf{Work.}}}
& \textbf{\textsc{Single-Agent}}          & 47.60 & 54.32 & 47.85 & 54.63 & 49.67 & 56.14 \\
& \textbf{\textsc{TIDE} w/o Templates}    & 62.48 & 64.35 & 65.88 & 67.92 & 72.74 & 75.04 \\
& \textbf{\textsc{TIDE} (Ours)}           & \textbf{69.06} & \textbf{70.46} & \textbf{67.64} & \textbf{68.76} & \textbf{76.53} & \textbf{77.82} \\
\midrule
\multirow{3}{*}{\rotatebox{90}{\textbf{Repo.}}}
& \textbf{\textsc{Single-Agent}}          & 8.66 & 10.34 & 11.15 & 12.92 & 12.19 & 13.27 \\
& \textbf{\textsc{TIDE} w/o Templates}    & 13.64 & 14.99 & 10.58 & 11.83 & 13.80 & 15.42 \\
& \textbf{\textsc{TIDE} (Ours)}           & \textbf{16.82} & \textbf{18.61} & \textbf{17.29} & \textbf{19.73} & \textbf{15.52} & \textbf{17.39} \\
\bottomrule
\end{tabular}
}
\vspace{-0.05in}
\caption{Contribution of each component of \textsc{TIDE} on both settings with GPT.}
\label{tab:ablation}
\vspace{-0.15in}
\end{table}

\paragraph{Effectiveness of Thought Templates.}
Having shown that iterative discovery enables \textsc{TIDE} to keep surfacing new problems across iterations, we now turn to what templates contribute.\rev{\footnote{The templates are distilled from held-out cases that share no project or problem with any test instance (please see Appendix~\ref{appen:disjoint} for details).}}
To this end, we ablate templates from \textsc{TIDE} and track retrieval coverage and precision, which measure \rev{how much of the gold pool is recovered and how often each predicted problem corresponds to a gold}, respectively.
Figure~\ref{fig:iter_recall_precision} shows that the template-guided variant yields an additional coverage gain over the no-template ablation (left) and a more pronounced precision margin at every iteration (right).
This shows that iteration and templates play complementary roles: iteration mainly drives how much of the gold pool is recovered, while templates mainly drive how accurate each recovered item is.
\rev{Table~\ref{tab:ablation} extends this ablation to all three sub-tasks.
Iteration alone already improves over \textsc{Single-Agent} on retrieval and resolution in both settings, and templates improve every component.}

\paragraph{Few-Shot as Template Substitute.}
We next ask whether few-shot demonstrations can replace the thought templates in \textsc{TIDE}.
To examine this, we follow the same iterative setup as \textsc{TIDE} but replace its thought templates with raw few-shot demonstrations drawn from the same training pool used to construct the templates.
As shown in Table~\ref{tab:fewshot}, demonstrations within the iterative loop fall well short of \textsc{TIDE} across retrieval, identification, and resolution.
This indicates that the value of templates lies in abstracting past examples into reusable reasoning patterns rather than in merely exposing the agent to them.

\begin{table}[t!]
\small
\centering
\renewcommand{\arraystretch}{1.05}
\renewcommand{\tabcolsep}{1.5mm}
\resizebox{\linewidth}{!}{%
\begin{tabular}{l cc cc cc}
\toprule
 & \multicolumn{2}{c}{\textbf{Retrieval}}
 & \multicolumn{2}{c}{\textbf{Identification}}
 & \multicolumn{2}{c}{\textbf{Resolution}} \\
\cmidrule(lr){2-3}\cmidrule(lr){4-5}\cmidrule(lr){6-7}
\textbf{Method} & \textbf{Cov.} & \textbf{F1} & \textbf{Cov.} & \textbf{F1} & \textbf{Cov.} & \textbf{F1} \\
\midrule
\midrule
\textbf{\textsc{Single-Agent}}   & 8.66  & 10.34 & 11.15 & 12.92 & 12.19 & 13.27 \\
\textbf{\textsc{Iter. + Demos}}  & 10.40 & 11.43 & 11.09 & 12.80 & 12.71 & 12.80 \\
\textbf{\textsc{TIDE} (Ours)}    & \textbf{16.82} & \textbf{18.61} & \textbf{17.29} & \textbf{19.73} & \textbf{15.52} & \textbf{17.39} \\
\bottomrule
\end{tabular}
}
\vspace{-0.05in}
\caption{Results on the Repository setting with GPT using raw few-shot demonstrations (\textsc{Iter. + Demos}).}
\label{tab:fewshot}
\vspace{-0.15in}
\end{table}

\newsavebox{\transferbox}
\savebox{\transferbox}{%
    \small
    \renewcommand{\arraystretch}{1.05}
    \renewcommand{\tabcolsep}{1.5mm}
    \resizebox{0.53\textwidth}{!}{%
    \begin{tabular}{l l cc cc cc}
    \toprule
     &
      & \multicolumn{2}{c}{\textbf{Retrieval}}
      & \multicolumn{2}{c}{\textbf{Identification}}
      & \multicolumn{2}{c}{\textbf{Resolution}} \\
    \cmidrule(lr){3-4}\cmidrule(lr){5-6}\cmidrule(lr){7-8}
    \textbf{Inference} & \textbf{Templates} & \textbf{Cov.} & \textbf{F1} & \textbf{Cov.} & \textbf{F1} & \textbf{Cov.} & \textbf{F1} \\
    \midrule
    \midrule
    \multirow{2}{*}{\textbf{GPT}}
      & GPT    & 16.82 & 18.61 & 17.29 & 19.73 & 15.52 & 17.39 \\
      & Gemini & 16.30 & 17.83 & 15.31 & 17.06 & 18.36 & 19.64 \\
    \noalign{\vskip 0.25ex}\cdashline{1-8}\noalign{\vskip 0.75ex}
    \multirow{2}{*}{\textbf{Gemini}}
      & Gemini & 22.55 & 25.14 & 21.93 & 24.22 & 24.32 & 26.98 \\
      & GPT    & 24.03 & 26.05 & 22.84 & 24.07 & 24.70 & 26.07 \\
    \bottomrule
    \end{tabular}
    }%
}
\begin{figure*}[t!]
    \centering
    \begin{minipage}[b]{0.45\linewidth}
        \centering
        \includegraphics[width=\linewidth,height=\dimexpr\ht\transferbox+\dp\transferbox\relax,keepaspectratio]{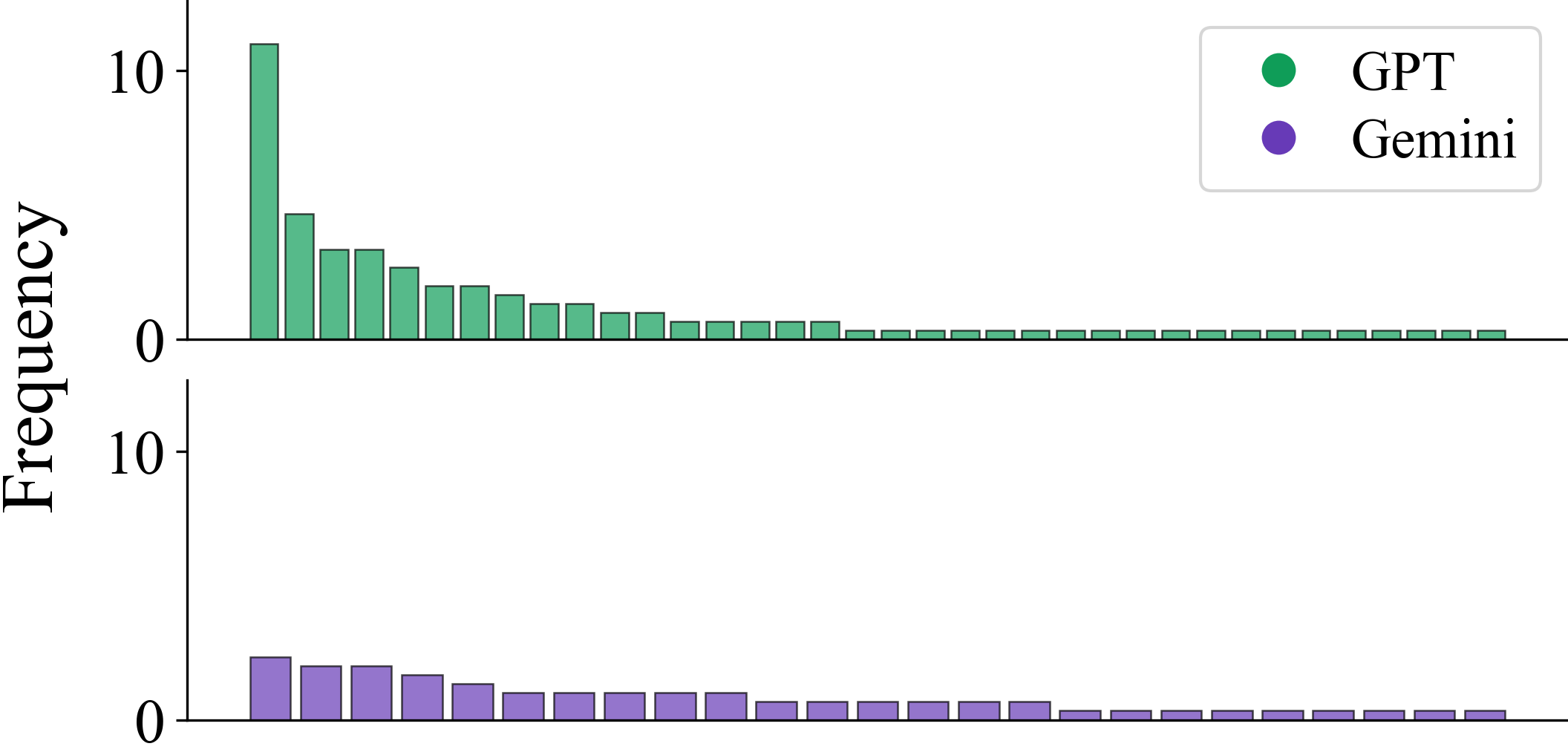}
        \captionof{figure}{Per-run template citation frequency.}\label{fig:template_frequency_gpt_vs_gem}
    \end{minipage}\hfill
    \begin{minipage}[b]{0.53\linewidth}
        \centering
        \usebox{\transferbox}
        \captionof{table}{Template transferability on the Repository setting.}\label{tab:transfer}
    \end{minipage}
\end{figure*}

\paragraph{Template Usage Distribution Across LLMs.}
Next, we examine how each backbone draws from its own template pool during inference.
Figure~\ref{fig:template_frequency_gpt_vs_gem} shows the average per-run usage frequency of cited templates, sorted in descending order for each LLM.
GPT concentrates more sharply on a few recurring templates, whereas Gemini spreads citations more evenly across its cited pool.
These divergent usage patterns raise the question: do templates built by one backbone still transfer to another?

\paragraph{Cross-LLM Template Transferability.}
To address this, we fix the inference LLM and vary the template source between GPT and Gemini.
As shown in Table~\ref{tab:transfer}, transferred templates perform comparably to self templates across all three components and in both directions, indicating that templates remain reusable across backbones despite each LLM's distinct usage profile.

\begin{figure}[t!]
    \centering
    \includegraphics[width=0.975\linewidth]{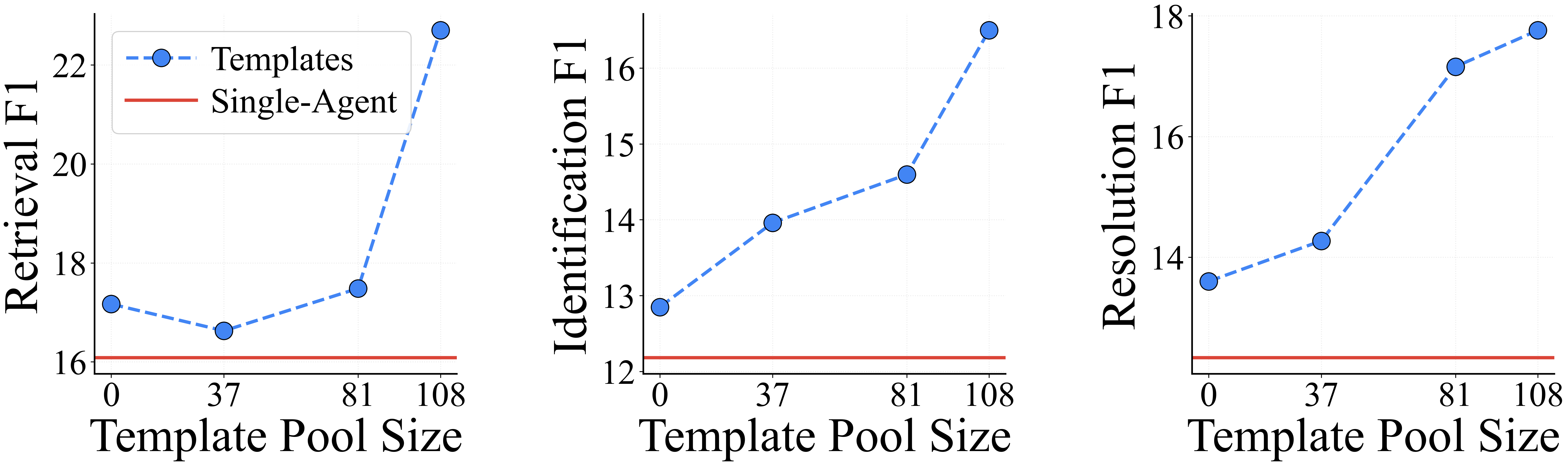}
    \vspace{-0.05in}
    \caption{F1 scores as the template pool size grows on the repository setting with Claude.}
    \label{fig:template_count_scaling}
    \vspace{-0.1in}
\end{figure}

\paragraph{Effect of Template Pool Size.}
To examine how performance scales with the size of the template pool, we vary the number of available templates and report results in Figure~\ref{fig:template_count_scaling}.
Notably, gains over \textsc{Single-Agent} grow with the pool size on all three components.
This indicates that each additional template supplies a problem class the agent can recognize, and that performance keeps improving as more solved cases are distilled into the pool.

\paragraph{Case Study.}
Beyond the quantitative gains shown above, we examine two representative cases, one from each evaluation setting, in Appendix~\ref{appen:case_study}.

\section{Related Work}

\paragraph{Task-oriented LLM Agents}
LLM agents have been increasingly studied in task-oriented environments that require document understanding~\cite{mmlongbench_doc}, tool use~\cite{Toolformer, ToolLLM}, web interaction~\cite{WebArena, Mind2Web}, or software engineering~\cite{SWE-agent, AutoCodeRover}, and evaluated on benchmarks that measure task completion in these environments~\cite{AgentBench, OSWorld}. These settings, however, typically presume a task already specified by a user request, issue description, failing test, or otherwise localized goal, leaving the agent reactive.

\paragraph{Proactive Agents}
Proactive agents aim to move beyond the reactive interaction model by anticipating user needs and initiating assistance before an explicit request is issued. One line of work focuses on uncovering user intent that goes beyond what is literally stated, either by asking clarification questions to resolve ambiguous requests~\cite{DBLP:conf/sigir/AliannejadiZCC19, Kuhn2022CLAMSC, DBLP:conf/emnlp/Zhang0RNC24, DBLP:journals/corr/abs-2511-02208, DiscoverLLM} or by navigating knowledge gaps that have not been articulated~\cite{proper}; these approaches, however, still presume a user-issued query as the anchor of interaction. A more recent line broadens the scope, studying when an agent should intervene~\cite{DBLP:conf/chi/LiuFSWIC25, DBLP:conf/emnlp/ZhangDLMKDCM25}, how user activity or signals can be used to anticipate assistance opportunities~\cite{ProactiveAgent, ContextAgent, FingerTip20K}, and how proactive suggestions should be generated and surfaced~\cite{probe}. Yet across this literature, proactivity remains anchored to a single localized need at a time, leaving open how an agent should jointly surface, ground, and resolve the many coexisting problems that real workflows typically contain.

\paragraph{Templates for LLM Reasoning}
Improving LLM reasoning has traditionally relied on internal model capability, whether by eliciting intermediate steps~\cite{cot, zeroshotcot} or by having a model critique and revise its own outputs~\cite{self-refine, reflexion}. More recent work observes that useful reasoning patterns recur across problems and externalizes them as reusable templates that can be retrieved and applied: Buffer-of-Thoughts~\cite{BoT} caches prior reasoning traces for retrieval on new problems, and follow-up work extends this idea to hierarchical template paths~\cite{ReasonFlux, SuperCorrect}, schema-based abstractions~\cite{SA-ICL}, self-evolving agent memory~\cite{ReasoningBank}, graph-based reuse of thought fragments~\cite{RoT}, and long-context multi-hop reasoning~\cite{ToTAL}. These template-based approaches, however, share a common assumption that the problem statement is already given and templates serve as schemas for how to solve it. We instead repurpose templates as discovery schemas that specify what contextual signals to attend to and how to connect them in order to infer a problem that has not been stated, and apply them iteratively so that each round extends coverage over coexisting problems rather than refines a single solution.

\section{Conclusion}
We presented \textsc{TIDE}, a framework for discovering multiple hidden problems from context through iterative discovery and thought templates.
Across personal workspaces and software repositories, \textsc{TIDE} consistently outperforms single-shot and multi-agent baselines on retrieval, identification, and resolution, with iteration and templates contributing complementary gains and templates transferring across backbones.
We believe these findings recast proactive assistance as a multi-step discovery process over context, offering a recipe for agents that surface what users would not have thought to ask.

\section*{Limitations}
Our \textsc{TIDE} delivers consistent gains across two realistic settings and four backbones, and the design choices behind it open a couple of interesting directions for further work.
First, templates are built once from a pool of solved cases and remain fixed at inference, which already proves effective and transfers across backbones; however, updating the library online from agent interactions, or augmenting the pool with automatically constructed cases, are natural extensions.
Likewise, iterative discovery trades a small bounded budget for broader coverage, a trade-off our analyses show is favorable against multi-agent baselines at matched budgets, and further investigating this iterative paradigm would be an exciting direction for future work.

\section*{Ethics Statement}
Our \textsc{TIDE} is designed to assist users by surfacing hidden problems from their working context that would otherwise go unaddressed, ranging from overlooked bottlenecks in personal workspaces to coexisting bugs in software repositories.
Since the framework operates over real-world documents and distills templates from previously solved cases, both of which may carry sensitive, biased, or otherwise undesirable content depending on the underlying source, we recommend applying standard safeguards such as content filtering, bias detection, and human-in-the-loop review at both template construction and deployment, in line with best practices for responsibly deploying LLM-based agents.

%\input{Sections/11_acknowledgement}

\bibliography{custom}

\appendix

\clearpage
\appendix

\begin{table*}[t!]
\small
\centering
\renewcommand{\arraystretch}{1.2}
\setlength{\tabcolsep}{3pt}
\resizebox{\linewidth}{!}{%
\begin{tabular}{l l rcr rcr rcr rcr}
\toprule
& & \multicolumn{3}{c}{\textbf{\textsc{TIDE} vs. \textsc{Single}}} & \multicolumn{3}{c}{\textbf{\textsc{TIDE} vs. \textsc{Multi}}} & \multicolumn{3}{c}{\textbf{\textsc{TIDE} vs. \textsc{Self-Refine}}} & \multicolumn{3}{c}{\textbf{\textsc{TIDE} vs. \textsc{Reflexion}}} \\
\cmidrule(lr){3-5} \cmidrule(lr){6-8} \cmidrule(lr){9-11} \cmidrule(lr){12-14}
& \textbf{Backbone} & $\Delta$ & \textbf{CI} & \textbf{$p$} & $\Delta$ & \textbf{CI} & \textbf{$p$} & $\Delta$ & \textbf{CI} & \textbf{$p$} & $\Delta$ & \textbf{CI} & \textbf{$p$} \\
\midrule
\midrule
% ================= Workspace =================
\multirow{4}{*}{\textbf{Workspace}}
& \textbf{GPT}     & $+$20.01 & [16.92, 23.04] & $<$0.001 & $+$35.92 & [32.80, 39.09] & $<$0.001 & $+$23.24 & [20.06, 26.34] & $<$0.001 & $+$20.97 & [18.61, 23.17] & $<$0.001 \\
& \textbf{Gemini}  & $+$21.94 & [17.19, 26.81] & $<$0.001 & $+$25.30 & [21.23, 29.24] & $<$0.001 & $+$9.84 & [5.97, 13.81] & $<$0.001 & $+$8.76 & [5.23, 12.20] & $<$0.001 \\
& \textbf{Claude}  & $+$20.42 & [15.10, 25.92] & $<$0.001 & $+$14.13 & [9.25, 19.27] & $<$0.001 & $+$2.96 & [$-$0.98, 7.13] & 0.112 & $+$4.66 & [$-$0.30, 9.86] & 0.059 \\
& \textbf{Qwen}    & $+$17.19 & [13.21, 21.57] & $<$0.001 & $+$14.13 & [9.35, 18.88] & $<$0.001 & $+$24.79 & [17.20, 32.22] & $<$0.001 & $+$15.60 & [11.03, 20.31] & $<$0.001 \\

\midrule
\midrule

% ================= Repository =================
\multirow{4}{*}{\textbf{Repository}}
& \textbf{GPT}     & $+$6.13 & [3.71, 8.73] & $<$0.001 & $+$6.22 & [1.89, 10.84] & 0.007 & $+$7.22 & [3.32, 11.69] & $<$0.001 & $+$6.84 & [2.91, 11.29] & $<$0.001 \\
& \textbf{Gemini}  & $+$8.47 & [4.00, 12.95] & $<$0.001 & $+$7.54 & [1.68, 13.38] & 0.018 & $+$19.46 & [10.98, 28.55] & $<$0.001 & $+$16.63 & [8.26, 25.43] & $<$0.001 \\
& \textbf{Claude}  & $+$5.68 & [1.67, 10.58] & 0.004 & $+$7.72 & [2.85, 13.22] & 0.002 & $+$6.29 & [1.98, 12.01] & 0.003 & $+$5.83 & [1.92, 10.47] & 0.003 \\
& \textbf{Qwen}    & $+$4.17 & [$-$0.13, 8.82] & 0.056 & $+$3.44 & [1.16, 5.88] & 0.005 & $+$2.68 & [0.31, 5.42] & 0.027 & $+$1.38 & [0.00, 2.92] & 0.050 \\
\bottomrule
\end{tabular}
}
\vspace{-0.05in}
\caption{Gains of \textsc{TIDE} over each baseline in Table~\ref{tab:main}, computed on the average across the six reported metrics. For each comparison we report the mean difference $\Delta$, its $90\%$ confidence interval, and the one-sided $p$-value, all from a paired bootstrap over instances, with the interval level matched to the one-sided test.}
\label{tab:significance}
%\vspace{-0.1in}
\end{table*}

\section{\texorpdfstring{\rev{Additional Implementation Details}}{Additional Implementation Details}}
\label{appen:impl}

\begin{revblock}
\paragraph{Decoding}
All methods run through the same inference code and the same decoding configuration for a given backbone, so that \textsc{TIDE} and the baselines differ only in the discovery procedure itself. We override no sampling parameter, leaving the temperature, the top-$p$, and every other decoding knob at the default of each provider. Since the provider APIs do not guarantee deterministic decoding even under a fixed seed, we do not fix one, and instead control for sampling variance by averaging every number we report in Table~\ref{tab:main} over three independent runs, testing the resulting margins in Appendix~\ref{appen:significance}. The LLM judge is held fixed across all methods and runs at the temperature of $1.0$ that the GPT-5 family requires, while the two additional judges of Appendix~\ref{appen:judge} run at a temperature of $0.1$.

\paragraph{Templates and Rounds}
Every template pool is distilled once offline and then kept fixed across all runs and analyses. The training cases behind the templates are produced by the same construction procedure as the evaluation instances of each setting, drawn from workspaces and repositories that are held out from the test split, namely $20$ training workspaces that carry one problem each and $25$ multi-bug training instances that carry $108$ problems in total. We distill one template per solved problem in those cases, so the size of a pool is not a value we choose but a reflection of how many problems its training source happens to contain. The number of rounds is likewise a budget cap rather than an estimate of how many problems an instance holds, since a run terminates as soon as a round returns nothing new. We also keep the identical full-context protocol for every backbone, so a workspace instance that exceeds a backbone's context window yields no prediction.
\end{revblock}

\section{\texorpdfstring{\rev{Significance of the Gains}}{Significance of the Gains}}
\label{appen:significance}

\begin{revblock}
We pair \textsc{TIDE} against each baseline instance by instance and compute one-sided $p$-values with a paired bootstrap, so that each reported interval describes how far a margin in Table~\ref{tab:main} could move had a different set of instances been drawn.

As shown in Table~\ref{tab:significance}, the gains of \textsc{TIDE} are significant at the $0.05$ level in most comparisons, across both settings and all four backbones.
\end{revblock}

\section{\texorpdfstring{\rev{Reliability of the LLM Judge}}{Reliability of the LLM Judge}}
\label{appen:judge}

\begin{table*}[t!]
\small
\centering
\renewcommand{\arraystretch}{1.05}
\setlength{\tabcolsep}{6pt}
{\footnotesize
\begin{tabular*}{\linewidth}{@{\extracolsep{\fill}} l l cc cc cc cc}
\toprule
& & \multicolumn{4}{c}{\textbf{Workspace}} & \multicolumn{4}{c}{\textbf{Repository}} \\
\cmidrule(lr){3-6} \cmidrule(lr){7-10}
& & \multicolumn{2}{c}{\textbf{Identification}} & \multicolumn{2}{c}{\textbf{Resolution}}
  & \multicolumn{2}{c}{\textbf{Identification}} & \multicolumn{2}{c}{\textbf{Resolution}} \\
\cmidrule(lr){3-4} \cmidrule(lr){5-6} \cmidrule(lr){7-8} \cmidrule(lr){9-10}
& \textbf{Methods}
  & \textbf{Cov.} & \textbf{F1} & \textbf{Cov.} & \textbf{F1}
  & \textbf{Cov.} & \textbf{F1} & \textbf{Cov.} & \textbf{F1} \\
\midrule
\midrule
% ================= DeepSeek V4 Pro judge =================
\multirow{5}{*}{\rotatebox{90}{\textbf{DeepSeek}}}
& \textbf{\textsc{Single-Agent}}  & 64.03 & 71.53 & 47.33 & 53.33 & 8.75 & 10.17 & 8.42 & 9.97 \\
& \textbf{\textsc{Multi-Agent}}   & 31.05 & 44.49 & 20.67 & 29.72 & 9.17 & 11.33 & 6.22 & 7.69 \\
& \textbf{\textsc{Self-Refine}}   & 50.92 & 61.77 & 38.61 & 45.88 & 6.67 & 7.50 & 3.17 & 3.50 \\
& \textbf{\textsc{Reflexion}}     & 61.21 & 69.82 & 45.50 & 51.44 & 9.58 & 10.67 & 7.13 & 8.13 \\
& \textbf{\textsc{TIDE} (Ours)}   & \textbf{82.55} & \textbf{84.10} & \textbf{64.56} & \textbf{65.65} & \textbf{11.67} & \textbf{13.05} & \textbf{9.25} & \textbf{10.57} \\

\midrule
\midrule

% ================= Gemini 3.5 Flash judge =================
\multirow{5}{*}{\rotatebox{90}{\textbf{Gemini}}}
& \textbf{\textsc{Single-Agent}}  & 60.12 & 67.20 & 61.67 & 69.14 & 12.43 & 13.22 & 8.92 & 10.57 \\
& \textbf{\textsc{Multi-Agent}}   & 29.68 & 42.58 & 27.20 & 39.37 & 11.96 & 14.31 & 7.88 & 9.49 \\
& \textbf{\textsc{Self-Refine}}   & 49.33 & 59.80 & 51.39 & 63.11 & 10.01 & 9.95 & 6.17 & 7.23 \\
& \textbf{\textsc{Reflexion}}     & 58.73 & 67.21 & 57.29 & 65.61 & 11.71 & 11.75 & 7.63 & 8.56 \\
& \textbf{\textsc{TIDE} (Ours)}   & \textbf{78.69} & \textbf{80.18} & \textbf{78.03} & \textbf{79.52} & \textbf{15.47} & \textbf{16.73} & \textbf{11.78} & \textbf{13.31} \\
\bottomrule
\end{tabular*}}
\vspace{-0.05in}
\caption{Effect of swapping the LLM judge, where rows are grouped by the judge that produced the scores, namely DeepSeek V4 Pro~\cite{deepseek2026v4} and Gemini 3.5 Flash~\cite{google2026gemini35flash}.}
\label{tab:judge_swap}
% \vspace{-0.1in}
\end{table*}

\begin{table}[t!]
\small
\centering
\renewcommand{\arraystretch}{1.05}
\renewcommand{\tabcolsep}{1.5mm}
\resizebox{\linewidth}{!}{%
\begin{tabular}{l l c cc cc}
\toprule
& & & \multicolumn{2}{c}{\textbf{Judge--Human}} & \multicolumn{2}{c}{\textbf{Human--Human}} \\
\cmidrule(lr){4-5} \cmidrule(lr){6-7}
& \textbf{Sub-task} & \textbf{\#}
  & \textbf{Exact} & \textbf{$\pm$1}
  & \textbf{Exact} & \textbf{$\kappa$} \\
\midrule
\midrule
\multirow{2}{*}{\textbf{Workspace}}
& \textbf{Identification} & 15 & 86.7 & 100.0 & 88.9 & 0.811 \\
& \textbf{Resolution}     & 15 & 93.3 & 100.0 & 86.7 & 0.775 \\
\cmidrule(lr){1-7}
\multirow{2}{*}{\textbf{Repository}}
& \textbf{Identification} & 15 & 95.0 & 100.0 & 92.2 & 0.880 \\
& \textbf{Resolution}     & 15 & 81.7 &  98.3 & 84.4 & 0.725 \\
\cmidrule(lr){1-7}
\multicolumn{2}{l}{\textbf{All}} & 60 & \textbf{89.2} & \textbf{99.6} & 88.1 & 0.802 \\
\bottomrule
\end{tabular}
}
\vspace{-0.05in}
\caption{Agreement between the LLM judge and human annotators over $60$ (gold, prediction) pairs.}
\label{tab:human_judge}
\vspace{-0.1in}
\end{table}

\begin{revblock}
\paragraph{Robustness to the Judge Model}
Since identification and resolution are open-ended, we score them with an LLM judge, following the practice of prior work in this setting~\cite{geval, probe}. To test how much the results depend on this choice, we re-score the identical stored predictions of the GPT backbone with two additional judges drawn from other model families, DeepSeek V4 Pro~\cite{deepseek2026v4}, which shares no family with any of the four evaluated backbones, and Gemini 3.5 Flash~\cite{google2026gemini35flash}. No re-inference is involved, and the pairing of gold problems with predictions is also left untouched, since matching uses the overlap between the predicted and gold-annotated evidence IDs and is therefore judge-independent.
As shown in Table~\ref{tab:judge_swap}, \textsc{TIDE} attains the best coverage and F1 on both judge-scored components, in both settings, and under either judge.

\paragraph{Agreement with Human Annotators}
Swapping the judge shows that the lead of \textsc{TIDE} does not depend on any single model, but it leaves open whether the scores themselves agree with human reading. To this end, we sample $60$ (gold, prediction) pairs from the stored evaluations, balanced across both settings and both sub-tasks, and spanning the low, middle, and high bands of the Likert scale. Each pair is labeled by four annotators, blind to the score of the judge, as correct, partially correct, or incorrect, giving $240$ human judgments from $12$ annotators. The annotators were recruited from graduate students in a computer science program, all fluent in English and able to read the repository items, and they took part voluntarily, with informed consent, and received approximately $30$ USD for their time. To compare the two, we map the score of the judge onto the same three levels, and report how often its label matches the human one exactly, how often the two fall within one level, and how often the annotators agree.

As shown in Table~\ref{tab:human_judge}, the judge and the annotators agree on most judgments, and land within one level on nearly all of them. The annotators agree with each other at a comparable rate, with a substantial inter-annotator agreement of $0.80$ in Fleiss $\kappa$~\cite{LandisKoch}, so the judge disagrees with the annotators no more than they disagree among themselves. The raw Likert scores also track the human labels, averaging $0.4$ for incorrect, $5.5$ for partially correct, and $9.1$ for correct, and correlating with them at a Spearman $\rho$ of $0.92$. Aggregating the four labels per pair into a majority label leaves the picture unchanged, with a Cohen's $\kappa$ of $0.87$ against the judge.
\end{revblock}

\section{\texorpdfstring{\rev{Project-Level Template Disjointness}}{Project-Level Template Disjointness}}
\label{appen:disjoint}

\begin{table}[t!]
\small
\centering
\renewcommand{\arraystretch}{1.05}
\renewcommand{\tabcolsep}{1.5mm}
\resizebox{\linewidth}{!}{%
\begin{tabular}{l cc cc cc}
\toprule
 & \multicolumn{2}{c}{\textbf{Retrieval}}
 & \multicolumn{2}{c}{\textbf{Identification}}
 & \multicolumn{2}{c}{\textbf{Resolution}} \\
\cmidrule(lr){2-3}\cmidrule(lr){4-5}\cmidrule(lr){6-7}
\textbf{Method} & \textbf{Cov.} & \textbf{F1} & \textbf{Cov.} & \textbf{F1} & \textbf{Cov.} & \textbf{F1} \\
\midrule
\midrule
\textbf{\textsc{Single-Agent}}   & 8.66  & 10.34 & 11.15 & 12.92 & 12.19 & 13.27 \\
\textbf{\textsc{Multi-Agent}}    & 10.11 & 12.66 & 10.19 & 12.77 & 9.89  & 12.39 \\
\midrule
\textbf{\textsc{TIDE} (Ours)}    & 16.82 & 18.61 & \textbf{17.29} & \textbf{19.73} & 15.52 & 17.39 \\
\textbf{\textsc{TIDE} (Project-Disjoint)} & \textbf{17.28} & \textbf{19.13} & 16.28 & 18.07 & \textbf{17.98} & \textbf{19.12} \\
\bottomrule
\end{tabular}
}
\vspace{-0.05in}
\caption{Results on the Repository setting with GPT, where the template pool is rebuilt so that no template shares a project with any test instance.}
\label{tab:disjoint}
\vspace{-0.1in}
\end{table}

\begin{revblock}
We construct the templates so that their source cases do not overlap with the test split, and we verify that separation here under a stricter, project-level criterion. In the workspace setting, the separation is complete, since the templates are distilled from training workspaces whose personas are disjoint from the test workspaces. In the repository setting, there are likewise no overlapping issues between the template sources and the test set, where eight of the eleven test projects do not appear in the template sources at all, and the remaining three appear only through entirely different issues at different commits. To make the separation exact, we rebuild the pool after dropping every template distilled from those three projects, so that no template shares a project with any test instance, and re-run the setting with it. As shown in Table~\ref{tab:disjoint}, performance stays well above both baselines and close to the original pool, which indicates that the templates carry patterns that transfer across projects.
\end{revblock}

\begin{table*}[t!]
\small
\centering
\renewcommand{\arraystretch}{1.15}
\setlength{\tabcolsep}{4pt}
\resizebox{\linewidth}{!}{%
\begin{tabular}{l cc cc cc cc cc cc}
\toprule
& \multicolumn{6}{c}{\textbf{Workspace}} & \multicolumn{6}{c}{\textbf{Repository}} \\
\cmidrule(lr){2-7} \cmidrule(lr){8-13}
& \multicolumn{2}{c}{\textbf{Retrieval}} & \multicolumn{2}{c}{\textbf{Identification}} & \multicolumn{2}{c}{\textbf{Resolution}}
& \multicolumn{2}{c}{\textbf{Retrieval}} & \multicolumn{2}{c}{\textbf{Identification}} & \multicolumn{2}{c}{\textbf{Resolution}} \\
\cmidrule(lr){2-3} \cmidrule(lr){4-5} \cmidrule(lr){6-7} \cmidrule(lr){8-9} \cmidrule(lr){10-11} \cmidrule(lr){12-13}
\textbf{Method}
  & \textbf{Cov.} & \textbf{F1} & \textbf{Cov.} & \textbf{F1} & \textbf{Cov.} & \textbf{F1}
  & \textbf{Cov.} & \textbf{F1} & \textbf{Cov.} & \textbf{F1} & \textbf{Cov.} & \textbf{F1} \\
\midrule
\midrule
\textbf{\textsc{Single-Agent}} & 47.60 & 54.32 & 47.85 & 54.63 & 49.67 & 56.14 & 8.66 & 10.34 & 11.15 & 12.92 & 12.19 & 13.27 \\
\midrule
\textbf{\textsc{Multi-Agent}} & 32.15 & 45.41 & 27.24 & 38.85 & 29.49 & 41.62 & 10.11 & 12.66 & 10.19 & 12.77 & 9.89 & 12.39 \\
\quad \textbf{+ Deduplication} & 28.45 & 40.64 & 26.60 & 37.94 & 28.70 & 40.84 & 9.04 & 11.27 & 10.15 & 12.63 & 9.87 & 11.83 \\
\quad \textbf{+ Diversity Prompting} & 38.04 & 50.60 & 33.56 & 45.30 & 36.42 & 47.99 & 8.12 & 10.67 & 9.02 & 10.75 & 12.49 & 12.64 \\
\midrule
\textbf{\textsc{TIDE} (Ours)} & \textbf{69.06} & \textbf{70.46} & \textbf{67.64} & \textbf{68.76} & \textbf{76.53} & \textbf{77.82} & \textbf{16.82} & \textbf{18.61} & \textbf{17.29} & \textbf{19.73} & \textbf{15.52} & \textbf{17.39} \\
\bottomrule
\end{tabular}
}
\vspace{-0.075in}
\caption{Results of the two stronger \textsc{Multi-Agent} variants on both settings with GPT, where deduplication merges predictions that describe the same problem and diversity prompting asks each agent to seek problems the others are likely to miss.}
\label{tab:multiagent_variants}
\vspace{-0.1in}
\end{table*}

\section{\texorpdfstring{\rev{Stronger Multi-Agent Variants}}{Stronger Multi-Agent Variants}}
\label{appen:variants}

\begin{revblock}
Since parallel agents share no state, they tend to re-discover the same salient problems, and we therefore ask whether a variant that targets this redundancy directly can close the gap. To examine this, we build two such variants on top of \textsc{Multi-Agent}. The first applies post-hoc deduplication, where an LLM groups the predictions that describe the same problem and keeps the best-grounded one from each group. The second applies diversity prompting, where every agent after the first looks for less salient problems that the others are likely to miss. As shown in Table~\ref{tab:multiagent_variants}, neither variant approaches \textsc{TIDE}. Deduplication does not help, since the parallel agents converge on only a couple of distinct problems per instance, so removing redundancy cannot recover a problem that no agent found. Diversity prompting raises coverage on the Workspace setting, yet it remains below even \textsc{Single-Agent}, and it lowers retrieval and identification on the Repository setting. Instructing an agent to avoid the obvious is therefore not a substitute for telling it what has already been found.
\end{revblock}

\section{\texorpdfstring{\rev{Token and Cost Accounting}}{Token and Cost Accounting}}
\label{appen:token}

\begin{table}[t!]
\small
\centering
\renewcommand{\arraystretch}{1.05}
\renewcommand{\tabcolsep}{1.5mm}
\resizebox{\linewidth}{!}{%
\begin{tabular}{l cccc}
\toprule
\textbf{Method} & \textbf{Calls} & \textbf{Input} & \textbf{Output} & \textbf{Cost} \\
\midrule
\midrule
\textbf{\textsc{Single-Agent}} & 1.0 & 0.16M & 3.2K & \$0.047 \\
\textbf{\textsc{Multi-Agent}}  & 10.0 & 1.63M & 17.8K & \$0.442 \\
\textbf{\textsc{TIDE} (Ours)}  & 9.2 & 1.54M & 29.2K & \$0.444 \\
\bottomrule
\end{tabular}
}
\vspace{-0.05in}
\caption{Measured LLM calls, tokens, and cost per instance on the Workspace setting with GPT.}
\label{tab:token}
\vspace{-0.15in}
\end{table}

\begin{revblock}
We match \textsc{TIDE} to \textsc{Multi-Agent} on the number of LLM calls to control the number of discovery opportunities, which leaves open how much each method spends within a call, since every round of \textsc{TIDE} additionally carries the cumulative discovery state and the template pool. To examine this, we measure the tokens that each method actually consumes, and price them at the list rate of the backbone. As shown in Table~\ref{tab:token}, the shared context $\mathcal{D}$ dominates every call, so the cumulative state and the template pool leave the input per call almost unchanged, and since \textsc{TIDE} terminates early once a round returns an empty batch, it issues fewer calls and consumes less total input per instance, at a comparable cost. Matching the two methods on calls therefore also matches them on tokens.
\end{revblock}

\section{Qualitative Study}
\label{appen:case_study}

We provide a qualitative analysis on two representative cases, one from each evaluation setting.
In the workspace case (Table~\ref{tab:case_study_workspace}), the gold issue is that the volunteer-tracking platform double-counts Mar 8 \textit{Community Build Day} check-ins, and the vendor patch is blocked behind a pending IT Security access approval ahead of a Mar 20 senior-leadership briefing. \textsc{Single-Agent} surfaces only an unrelated facility-rider procurement stall and retrieves none of the gold documents, so the identification, the chosen action, and the addressee are all wrong. \textsc{TIDE}, by contrast, reaches the data-integrity issue in a later iteration, retrieves the gold documents, and escalates to the right manager with the gating access ticket, the vendor-deployment deadline, and the presentation deadline.
\rev{In the repository case (Table~\ref{tab:case_study_repo}), the gold issue is a callable-metric bug in \textit{mlxtend}'s permutation feature-importance utility. The function documents its \texttt{metric} argument as either a string or a callable, but it binds the scoring function only in the string branches, so a callable metric passes validation and then raises an \texttt{UnboundLocalError}. \textsc{Self-Refine} rewrites its initial set over the same number of rounds and still returns a single bottleneck about a different file, so the callable-metric bug is never surfaced. \textsc{TIDE} retrieves the function, names the missing fallback branch, and adds the assignment that the gold patch adds. The template that guided this prediction was mined from \textit{sympy}, where the same shape of defect appears in symbolic evaluation, so the retrieved abstraction transfers across repositories rather than restating a problem that the pool already contains.}
In both cases, \textsc{TIDE}'s prediction is guided by a thought template that captures a pattern recurring across instances within the same setting.

\section{Prompts}
\label{appen:prompts}

We list the four prompts used end-to-end in our pipeline: template construction (Figures~\ref{fig:prompt_template_construction_workspace} and~\ref{fig:prompt_template_construction_code}) and per-iteration inference (Figures~\ref{fig:prompt_inference_workspace} and~\ref{fig:prompt_inference_code}), each instantiated separately for the workspace and code settings. Placeholders in \texttt{\{curly braces\}} are filled at runtime; the \emph{previously found bottlenecks} block is omitted at iteration~1 and re-injected on subsequent iterations.

\clearpage
\begin{table*}[t!]
\centering
\renewcommand{\arraystretch}{1.15}
\setlength{\tabcolsep}{4pt}
\small
\begin{tabular}{@{}p{0.09\linewidth} p{0.21\linewidth} p{0.31\linewidth} p{0.33\linewidth}@{}}
\toprule
& \textbf{Retrieval} & \textbf{Identification} & \textbf{Resolution} \\
\midrule
\multicolumn{4}{@{}p{\linewidth}@{}}{\cellcolor{gray!10}\textbf{Workspace.}\ A corporate community-impact manager's workspace; the volunteer-tracking platform double-counts Mar 8 \textit{Community Build Day} check-ins, and the vendor's fix is blocked behind a pending IT Security access approval ahead of the Mar 20 senior leadership briefing.} \\
\midrule
\textbf{Gold}
& 5 supporting documents: the Mar 12 vendor support ticket from Emily Zhang, the Mar 13 AccessHub access-review request AR-2025-4411, the duplicated VolunteerHub records for the Mar 8 \textit{Community Build Day}, the Mar 20 senior-leadership pre-read deck, and the coordination thread on the data freeze.
& VolunteerHub double-counts Mar 8 check-ins because of a sync conflict between SSO and the manual check-in override, so reported volunteer hours are inflated; the vendor patch is staged but cannot deploy until IT Security signs off on AR-2025-4411, and uncorrected figures would land in the Mar 20 senior-leadership briefing.
& \texttt{escalate\_to\_manager} to David Chen, summarizing the duplication, the gating AccessHub ticket, the Mar 17 data freeze, and the Mar 20 readout so he can unblock IT Security in time for the vendor patch.
\\
\midrule
\textbf{\textsc{Single-Agent}}
& \textcolor{red!70!black}{$\times$}\ \textbf{0 of 5} gold documents (returns only a procurement-intake stall on the Apr 10 Showcase facility rider and a calendar double-booking, neither touching the VolunteerHub problem).
& \textcolor{red!70!black}{$\times$}\ Surfaces an unrelated facility-rider procurement intake that has been sitting unassigned since Feb 28; the VolunteerHub data-integrity issue is not surfaced at all.
& \textcolor{red!70!black}{$\times$}\ \texttt{send\_email} to the procurement intake about the facility rider; the action does not target David Chen, the AccessHub ticket, or the Mar 20 metrics deadline.
\\
\midrule
\textbf{\textsc{TIDE} (Ours)}
& \textcolor{green!50!black}{$\checkmark$}\ \textbf{5 of 5} gold documents, no spurious documents. (iteration 3)
& \textcolor{green!50!black}{$\checkmark$}\ Names the VolunteerHub Mar 8 duplication, ties it to the AccessHub ticket AR-2025-4411 stalled in Security, and links the chain to the Mar 20 senior-leadership briefing.
& \textcolor{green!50!black}{$\checkmark$}\ \texttt{escalate\_to\_manager} to David Chen with a sharp summary of the inflation, the gating Security approval, the Mar 18 vendor deployment window, and the Mar 20 readout.
\\
\midrule
\multicolumn{4}{@{}p{\linewidth}@{}}{\cellcolor{blue!4}\footnotesize \textbf{\textsc{TIDE} template used:}\ \texttt{[TID\_11]\ Unblocking High-Stakes Deliverable Under SME Communication Lag}. \emph{Evidence flow (first three steps):} Identify a high-stakes, time-bounded deliverable review deadline with an external stakeholder. Verify if the current deliverable iteration relies on outdated, legacy metrics or placeholder files. Confirm that a newer, finalized source asset has been uploaded/delivered by a specialized contributor.} \\
\bottomrule
\end{tabular}
\vspace{-0.05in}
\caption{Case study from the personal-workspace setting (a corporate community-impact manager whose volunteer-tracking platform double-counts \textit{Community Build Day} check-ins ahead of a senior-leadership briefing). Rows are the three methods (Gold, \textsc{Single-Agent}, \textsc{TIDE}) and columns are the three sub-tasks (retrieval, identification, resolution). \textcolor{green!50!black}{$\checkmark$} marks a sub-task the method handles correctly; \textcolor{red!70!black}{$\times$} marks a wrong or missing one. The shaded row at the bottom shows the thought template that \textsc{TIDE} retrieved for this prediction.}
\label{tab:case_study_workspace}
\end{table*}

\clearpage
\begin{table*}[t!]
\centering
\renewcommand{\arraystretch}{1.15}
\setlength{\tabcolsep}{4pt}
\small
\begin{tabular}{@{}p{0.09\linewidth} p{0.21\linewidth} p{0.31\linewidth} p{0.33\linewidth}@{}}
\toprule
& \textbf{Retrieval} & \textbf{Identification} & \textbf{Resolution} \\
\midrule
\multicolumn{4}{@{}p{\linewidth}@{}}{\cellcolor{gray!10}\textbf{Repository.}\ \textit{mlxtend}, an instance whose gold pool holds three independent problems. The one shown here lives in \texttt{mlxtend/\allowbreak evaluate/\allowbreak feature\_\allowbreak importance.py}, where \texttt{feature\_\allowbreak importance\_\allowbreak permutation} documents \texttt{metric} as a string or a callable but binds \texttt{score\_func} only in the string branches, so a callable metric passes validation and then fails at call time.} \\
\midrule
\textbf{Gold}
& 1 gold buggy function, \texttt{feature\_\allowbreak importance\_\allowbreak permutation} in \texttt{mlxtend/\allowbreak evaluate/\allowbreak feature\_\allowbreak importance.py}.
& \texttt{score\_func} is assigned inside \texttt{if metric == 'r2'} and \texttt{elif metric == 'accuracy'} with no fallback branch, so a custom callable such as \texttt{mean\_\allowbreak squared\_\allowbreak error} leaves it unbound and raises \texttt{Unbound\allowbreak LocalError} when the baseline score is computed.
& Add the missing fallback so that a callable metric is used directly, \texttt{else:\allowbreak{} score\_func = metric}.
\\
\midrule
\textbf{\textsc{Self-Refine}}
& \textcolor{red!70!black}{$\times$}\ After its refinement rounds it returns a single bottleneck pointing at \texttt{mcnemar\_tables}; \texttt{feature\_\allowbreak importance\_\allowbreak permutation} is never retrieved.
& \textcolor{red!70!black}{$\times$}\ Reports that \texttt{combinations} is never imported in \texttt{mcnemar.py} and would raise \texttt{NameError}, which is none of the three gold problems.
& \textcolor{red!70!black}{$\times$}\ Emits a one-line patch adding \texttt{from itertools import combinations}, an edit that no gold patch touches.
\\
\midrule
\textbf{\textsc{TIDE} (Ours)}
& \textcolor{green!50!black}{$\checkmark$}\ Retrieves exactly \texttt{feature\_\allowbreak importance\_\allowbreak permutation}, with no spurious documents.
& \textcolor{green!50!black}{$\checkmark$}\ Names the non-exhaustive branch assignment, ties it to the documented callable-metric contract, and locates the failure at the baseline-score call.
& \textcolor{green!50!black}{$\checkmark$}\ Adds \texttt{else:\allowbreak{} score\_func = metric} to the branch chain, matching the gold patch.
\\
\midrule
\multicolumn{4}{@{}p{\linewidth}@{}}{\cellcolor{blue!4}\footnotesize \textbf{\textsc{TIDE} template used:}\ \texttt{[TID\_81]\ Unbound\_\allowbreak Variable\_\allowbreak From\_\allowbreak Non\_\allowbreak Exhaustive\_\allowbreak Conditional\_\allowbreak Assignment}, mined from \textit{sympy}. \emph{Pattern:} A local variable is initialized within conditional branches (such as if/elif) without an exhaustive fallback (else). If none of the conditions are met, subsequent execution referencing the variable raises an UnboundLocalError.} \\
\bottomrule
\end{tabular}
\vspace{-0.05in}
\caption{Case study from the software-repository setting (\textit{mlxtend}, a callable-metric bug in the permutation feature-importance utility). Rows are the three methods (Gold, \textsc{Self-Refine}, \textsc{TIDE}) and columns are the three sub-tasks (retrieval, identification, resolution). \textcolor{green!50!black}{$\checkmark$} marks a sub-task the method handles correctly; \textcolor{red!70!black}{$\times$} marks a wrong or missing one. The shaded row at the bottom shows the thought template that \textsc{TIDE} retrieved for this prediction.}
\label{tab:case_study_repo}
\end{table*}

\clearpage
\begin{figure*}[t!]
\centering
\begin{tcolorbox}[colback=gray!5, colframe=gray!50,
                  boxrule=0.5pt, arc=2mm,
                  left=4pt, right=4pt, top=3pt, bottom=3pt]
\small
\textbf{Template Construction Prompt (Workspace).}

You are an expert at extracting reusable reasoning patterns from solved examples.

Given a solved bottleneck detection example, extract an abstract reasoning template.

\medskip
\textbf{Solved example shown to the model:}
\begin{verbatim}
Bottleneck: {bottleneck_description}

Evidence Documents Used:
- [{doc.type}]
  {doc.payload as JSON}
...

Checklist Steps:
- Retrieval:       {retrieval_description}
- Identification:  {identification_description}
- Task Execution:  {task_description}
\end{verbatim}

\medskip
\textbf{Rules:}
\begin{enumerate}[topsep=0pt, itemsep=0ex, parsep=0pt, left=14pt]
  \item \textbf{Domain-agnostic}: use general workplace language only; replace specific domain terms with role-based descriptions.
  \item Be \textbf{concise}.
  \item \textbf{Aggressive abstraction}: the template must apply to many scenarios beyond the solved example.
  \begin{itemize}[topsep=0pt, itemsep=0ex, parsep=0pt, left=12pt]
     \item Replace specific artifacts (e.g., ``allocation spreadsheet'', ``quarterly slide deck'') with generic types (e.g., ``a shared source artifact'', ``a deliverable'').
     \item Replace specific people, teams, or titles (e.g., ``analytics lead'', ``VP of Marketing'') with abstract role descriptors (e.g., ``the content owner'', ``the approver'').
     \item Replace specific events, dates, or deadlines with abstract time-pressure descriptors (e.g., ``an imminent time-bounded review'').
     \item Replace industry-specific jargon, tool names, or systems with generic terms (e.g., ``a documentation page'', ``a tooling capability'').
     \item Replace specific metrics or values with abstract terms (e.g., ``inconsistent key figures'', ``a material differential'').
  \end{itemize}
  \item \textbf{Preserve structural elements}: keep elements that define the pattern's identity (conflict type, time pressure, role dependencies, observable problem state); abstraction strips domain specifics, not structure.
  \item \textbf{The pattern must be testable against varied scenarios}: if a different domain with the same structural bottleneck would not match, abstract further.
\end{enumerate}

\medskip
\textbf{Output format:}
\begin{verbatim}
{
  "template_name": "Short descriptive name for this pattern",
  "pattern":       "Brief description of the bottleneck situation",
  "evidence_flow": ["What to check first",
                    "What to check next", ...]
}
\end{verbatim}

\textbf{Instruction constraint:} respond only with the JSON object.
\end{tcolorbox}
\caption{Template construction prompt for the workspace setting. Given one solved example, the model returns a reusable template added to the template pool $\mathcal{T}$. Only \texttt{pattern} and \texttt{evidence\_flow} are exposed to the agent at inference.}
\label{fig:prompt_template_construction_workspace}
\end{figure*}

\clearpage
\begin{figure*}[t!]
\centering
\begin{tcolorbox}[colback=gray!5, colframe=gray!50,
                  boxrule=0.5pt, arc=2mm,
                  left=4pt, right=4pt, top=3pt, bottom=3pt]
\small
\textbf{Template Construction Prompt (Code).}

You extract reusable, repo-agnostic bug patterns from solved code-bug examples.

The gold patch defines correct behavior: the pre-patch code is the bug shape, the post-patch code is the fix shape, and the diff encodes the intent.

\medskip
\textbf{Solved example shown to the model:}
\begin{verbatim}
Repository: {repo}

Issue / Bug Report:
{bottleneck_description}

Buggy Function(s):
### {func.qualname} (file: {func.path})
```python
{func.content}
```
...

Gold Patch (the fix):
```diff
{patch}
```
\end{verbatim}

\medskip
\textbf{Rules:}
\begin{enumerate}[topsep=0pt, itemsep=0ex, parsep=0pt, left=14pt]
  \item \textbf{Pattern and evidence\_flow: abstract}. Replace specific identifiers with role descriptors. Both fields must transfer to a different Python codebase carrying the same \emph{structural} bug.
  \item \textbf{Code-centric detection}: each step in \texttt{evidence\_flow} must reference observable code signals visible in the \emph{buggy code alone}. The gold patch is a reference for understanding the bug, but \texttt{evidence\_flow} must work without seeing the patch.
  \item \textbf{One pattern per template}.
  \item \textbf{Broad applicability}: pattern and evidence\_flow must describe a bug shape that could plausibly appear in multiple scenarios (e.g., parsers, serializers, validators, builders, schedulers). Aim for patterns where at least three different bug scenarios outside this example would still match.
\end{enumerate}

\medskip
\textbf{Output format:}
\begin{verbatim}
{
  "template_name": "Short descriptive name (abstract)",
  "pattern":       "Abstract: what shape of code carries the bug
                    and what input/condition triggers it.",
  "evidence_flow": ["Abstract: what to check first when scanning
                     a function for this pattern",
                    "Abstract: what to check next", ...]
}
\end{verbatim}

\textbf{Instruction constraint:} respond with the JSON object only.
\end{tcolorbox}
\caption{Template construction prompt for the code setting. Given one solved bug-fix example, the model returns a reusable bug-pattern template added to the template pool $\mathcal{T}$. Only \texttt{pattern} and \texttt{evidence\_flow} are exposed to the agent at inference.}
\label{fig:prompt_template_construction_code}
\end{figure*}
\clearpage
\begin{figure*}[t!]
\centering
\begin{tcolorbox}[colback=gray!5, colframe=gray!50,
                  boxrule=0.5pt, arc=2mm,
                  left=4pt, right=4pt, top=3pt, bottom=3pt]
\small
\textbf{Inference Prompt (Workspace).}

\medskip
\textbf{Role definition.} You are a proactive agent whose goal is to serve a user, about whom you will receive information in the ``World Model'' section. You will be provided personal context about the user, their relationships, their organizational roles and structures, their personal pain points and priorities, and the actions they are able to take. Your task is to find and identify (without prior prompting) a singular bottleneck in their lives, and recommend a resolution by selecting one of the pre-specified actions. The bottleneck must be identified by analyzing the provided documents, emails, and calendar events.

\medskip
\textbf{Thought templates definition.} The following reasoning templates describe common bottleneck patterns. Use these as guides when analyzing documents: match observed situations to relevant templates to select the right action and fill parameters correctly. Not every template will be relevant.

\medskip
\textbf{Bottleneck definition.} A bottleneck is a concerning issue that
\begin{itemize}[topsep=0pt, itemsep=0ex, parsep=0pt, left=12pt]
   \item represents an important pattern or situation requiring attention (based on the World Model);
   \item is directly resolvable through one of the provided user actions;
   \item would be flagged by a competent assistant as needing intervention;
   \item is directly indicated within a document or set of documents in the datastore.
\end{itemize}

\medskip
\textbf{Previously found bottlenecks} (included only after the first iteration). The following bottlenecks have already been identified in previous iterations. Do not repeat these; find only a new bottleneck that is different from those listed. If no more new bottlenecks remain, return an empty JSON object \texttt{\{\}}.

\medskip
\textbf{Task execution.}
\begin{enumerate}[topsep=0pt, itemsep=0ex, parsep=0pt, left=14pt]
   \item Analysis: examine all provided documents, emails, and calendar events to understand the user's current situation.
   \item Pattern recognition: identify concerning patterns that match the bottleneck definition.
   \item Action selection: choose the most appropriate action from the available actions list.
   \item Response generation: provide the analysis and recommendation in the specified JSON format.
\end{enumerate}

\medskip
\textbf{Output format.}
\begin{verbatim}
{
  "bottlenecks": [
    {
      "used_template_id":   "TID_X",
      "used_template_name": "...",
      "why_matched": "Which observations in the documents satisfy
                      the matched template's pattern, with doc IDs.",
      "retrieved_documents": ["doc_id_1", "doc_id_2", ...],
      "bottleneck": "Natural-language description of the issue.",
      "action": {
        "function_name": "action_function_name",
        "parameters":    {"parameter_name1": "value1", ...}
      }
    }
  ]
}
\end{verbatim}
The \texttt{bottlenecks} array contains exactly one identified bottleneck per iteration.

\medskip
\textbf{Context shown to the model:}
\begin{verbatim}
Persona:     {persona}
World Model: {world_model}

Bottleneck Pattern Templates:
[{template_id}] {template_name}
  Pattern: {pattern}
  Evidence flow:
    - {step_1}
    - {step_2}
    ...

Documents:         {data_sources}
Available Actions: {available_actions}
\end{verbatim}
\end{tcolorbox}
\caption{Per-iteration inference prompt for the workspace setting. At each iteration the model receives the persona, world model, retrieved documents, available action library, current template pool, and the bottlenecks already returned in previous iterations; it returns one new bottleneck and the chosen action, or an empty object to terminate.}
\label{fig:prompt_inference_workspace}
\end{figure*}
\clearpage
\begin{figure*}[t!]
\centering
\begin{tcolorbox}[colback=gray!5, colframe=gray!50,
                  boxrule=0.5pt, arc=2mm,
                  left=4pt, right=4pt, top=3pt, bottom=3pt]
\small
\textbf{Inference Prompt (Code).}

\medskip
\textbf{Role definition.} You are a proactive code-maintenance agent whose goal is to surface issues in a Python software repository. Your task is to find and identify (without prior prompting) distinct issues worth reporting in the codebase, and recommend a resolution for each by producing a unified diff patch.

\medskip
\textbf{Bug pattern templates definition.} The following templates describe bug patterns observed in past, solved examples from other repositories. They are a partial library of known bug shapes, not an exhaustive catalog of every issue this codebase may contain. Report two kinds of issues:
\begin{enumerate}[topsep=0pt, itemsep=0ex, parsep=0pt, left=14pt]
   \item Issues that match one of the templates below. Set \texttt{used\_template\_id} to the template id and explain the match in \texttt{why\_matched}.
   \item Issues that do not fit any template but are still genuine bugs grounded in the code. Leave \texttt{used\_template\_id} as \texttt{""}.
\end{enumerate}
Both kinds count equally.

\medskip
\textbf{Bottleneck definition.} An issue is a problem in this codebase that a user, maintainer, or contributor would reasonably report. Each issue must
\begin{itemize}[topsep=0pt, itemsep=0ex, parsep=0pt, left=12pt]
   \item be resolvable by editing one or more of the provided Python functions;
   \item point to specific function(s) and a specific change;
   \item be supported by concrete evidence drawn from the code itself;
   \item be distinct from other issues in the list (do not split one bug across multiple entries).
\end{itemize}

\medskip
\textbf{Previously found bottlenecks} (included only after the first iteration). Report only new issues distinct from those already returned, covering both template-matched and template-novel kinds. If no new issues remain, return an empty JSON object \texttt{\{\}}.

\medskip
\textbf{Task execution.}
\begin{enumerate}[topsep=0pt, itemsep=0ex, parsep=0pt, left=14pt]
   \item Analysis: examine all provided functions to understand what each one is supposed to do.
   \item Issue identification: identify problems matching the issue definition, skipping any already reported.
   \item Patch generation: produce a unified diff patch resolving each identified issue.
   \item Response generation: provide analysis and patches in the specified JSON format.
\end{enumerate}

\medskip
\textbf{Output format.}
\begin{verbatim}
{
  "bottlenecks": [
    {
      "used_template_id":   "TID_X" or "",
      "used_template_name": "..."   or "",
      "why_matched": "Which observations in the functions satisfy
                      the matched template's evidence_flow, citing
                      specific function IDs and lines. Use \"\"
                      if no template applies.",
      "retrieved_documents": ["func_id_1", "func_id_2", ...],
      "bottleneck": "Natural-language description, citing concrete
                     code evidence and qualnames.",
      "action": {
        "patch": "<unified diff text fixing the issue,
                  starting with `diff --git a/<path> b/<path>`>"
      }
    }
  ]
}
\end{verbatim}

\medskip
\textbf{Context shown to the model:}
\begin{verbatim}
Repository: {repo}

Bug Pattern Templates:
[{template_id}] {template_name}
  Pattern: {pattern}
  Evidence flow:
    - {step_1}
    - {step_2}
    ...

Functions: {data_sources}
\end{verbatim}
\end{tcolorbox}
\caption{Per-iteration inference prompt for the code setting. At each iteration the model receives the repository name, current template pool, candidate Python functions, and the bottlenecks already returned in previous iterations; it returns new bottlenecks (template-matched or template-novel) with unified-diff patches, or an empty object to terminate.}
\label{fig:prompt_inference_code}
\end{figure*}

\clearpage

\end{document}